\documentclass{article}

\usepackage{microtype}
\usepackage{graphicx}
\usepackage{subfigure}
\usepackage{booktabs} 

\usepackage{url}
\usepackage{color}
\usepackage{graphicx}
\usepackage{cuted}
\usepackage{capt-of}
\usepackage{tablefootnote}
\usepackage[font=small]{caption} 
\usepackage{booktabs} 
\usepackage{multirow} 
\usepackage{footnote}
\usepackage{amsmath}
\usepackage{enumitem}
\usepackage{adjustbox}
\usepackage{caption}
\usepackage{pifont}

\usepackage{hyperref}
\usepackage{sidecap}
\usepackage{natbib}

\newcommand\ignore[1]{} 

\usepackage[accepted]{icml2021}

\icmltitlerunning{Compositional Video Synthesis with Action Graphs}

\usepackage{amsmath,amsfonts,bm}

\def\Figref#1{Figure~\ref{#1}}

\def\secref#1{section~\ref{#1}}
\def\Secref#1{Section~\ref{#1}}

\def\eqref#1{equation~\ref{#1}}

\def\1{\bm{1}}

\DeclareMathAlphabet{\mathsfit}{\encodingdefault}{\sfdefault}{m}{sl}
\SetMathAlphabet{\mathsfit}{bold}{\encodingdefault}{\sfdefault}{bx}{n}

\hypersetup{
    colorlinks=true,
    linkcolor=blue,
    filecolor=blue,      
    urlcolor=blue,
}
\definecolor{lightgray}{gray}{0.9}
\definecolor{lightblue}{rgb}{0.93,0.95,1.0}
\definecolor{darkgreen}{rgb}{0.0,0.6,0.0}
\definecolor{darkblue}{rgb}{0.0,0.0,0.5}

\newcommand{\reals}{\mathbb{R}}

\newcommand{\cC}{\mathcal{C}}
\newcommand{\cR}{\mathcal{R}}

\newcommand{\cA}{\mathcal{A}}

\newcommand{\zz}{\boldsymbol{z}}
\newcommand{\uu}{\boldsymbol{u}}
\newcommand{\ag}{A}

\newcommand{\phiv}{\boldsymbol{\phi}}
\newcommand{\psiv}{\boldsymbol{\psi}}
\newcommand{\ignorebig}[1]{}
\newcommand{\tabref}[1]{Table~\ref{#1}}

\newcommand{\figgref}[1]{Fig.~\ref{#1}}

\begin{document}
\twocolumn[
\icmltitle{Compositional Video Synthesis with Action Graphs}

\icmlsetsymbol{equal}{*}

\begin{icmlauthorlist}
\icmlauthor{Amir Bar}{equal,to}
\icmlauthor{Roei Herzig}{equal,to}
\icmlauthor{Xiaolong Wang}{ucsd,ber}
\icmlauthor{Anna Rohrbach}{ber}
\icmlauthor{Gal Chechik}{nv,biu}\\
\icmlauthor{Trevor Darrell}{ber}
\icmlauthor{Amir Globerson}{to}
\end{icmlauthorlist}

\icmlaffiliation{to}{The Blavatnik School of Computer Science, Tel Aviv University}
\icmlaffiliation{ber}{UC Berkeley}
\icmlaffiliation{ucsd}{UC San Diego}
\icmlaffiliation{nv}{NVIDIA Research}
\icmlaffiliation{biu}{Bar-Ilan University}

\icmlcorrespondingauthor{Amir Bar}{amir.bar@cs.tau.ac.il}

\icmlkeywords{Machine Learning, ICML}

\vskip 0.3in
]

\printAffiliationsAndNotice{\icmlEqualContribution} 

\begin{abstract}
Videos of actions are complex signals containing rich compositional structure in space and time. Current video generation methods lack the ability to condition the generation on multiple coordinated and potentially simultaneous timed actions. To address this challenge, we propose to represent the actions in a graph structure called Action Graph and present the new ``Action Graph To Video'' synthesis task. Our generative model for this task (AG2Vid) disentangles motion and appearance features, and by incorporating a scheduling mechanism for actions facilitates a timely and coordinated video generation. We train and evaluate AG2Vid on the CATER and Something-Something V2 datasets, and show that the resulting videos have better visual quality and semantic consistency compared to baselines. Finally, our model demonstrates zero-shot abilities by synthesizing novel compositions of the learned actions.\footnote{See the project page for code and pretrained models:\newline~\url{https://roeiherz.github.io/AG2Video}.} 

\end{abstract}


\section{Introduction}
\label{sec:introduction}
\begin{figure}[t!]
\centering
    \includegraphics[width=.95\linewidth]{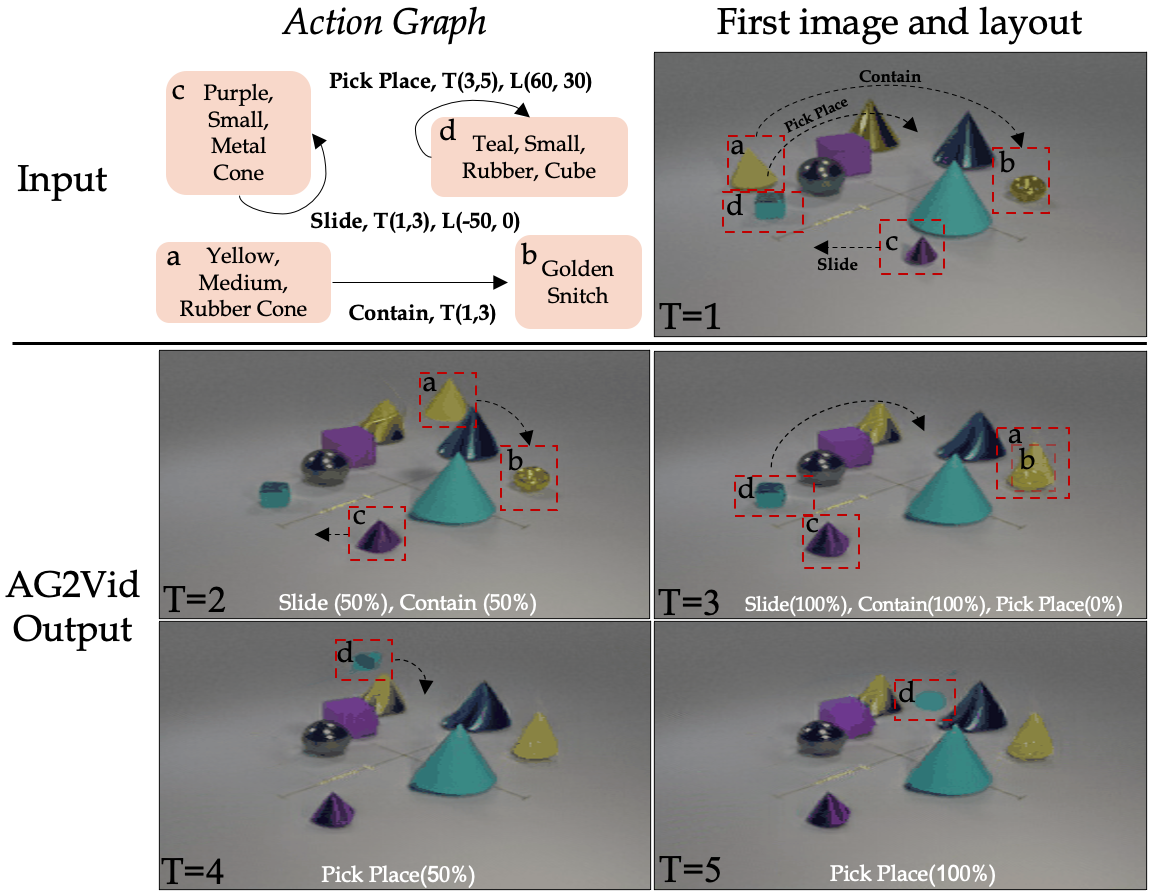}
    \captionof{figure}{We propose a new task called \textit{Action Graph to Video}. To represent input actions, we use a graph structure called \textit{Action Graph}, where nodes are objects and edges are actions with start/end frames specified by $T(s,e)$ and optional attributes (such as object destination coordinates) specified by $L(x,y)$, where $(x,y)\in \reals^2$ is a spatial offset. Together with input first frame and scene layout, our goal is to synthesize a video that follows the input Action Graph instructions. We include an example over CATER~\cite{girdhar2020cater} dataset. The annotated object-action trajectories are only shown for illustration purposes.}
\label{fig:teaser}
\end{figure}

Learning to generate visual content is a fundamental task in computer vision and graphics, with numerous applications from sim-to-real training of autonomous agents, to creating visuals for games and movies. While the quality of generating still images has leaped forward recently \citep{brock2018biggan,Karras2019stylegan2}, generating videos remains in its infancy. Generating  actions and interactions between objects is perhaps the most challenging aspect of conditional video generation.
When constructing a complex scene with multiple agents (e.g. a driving scene with several cars or a basketball game with a set of players) it would be important to have {\it control} over multiple agents and their actions as well as to {\it coordinate} them w.r.t. each other (e.g. to generate a Pick 'n' Roll move in basketball).
Actions create long-range spatio-temporal dependencies between people and objects they interact with. For example, when a player passes a ball, the entire movement sequence of all the entities (thrower, ball, receiver) must be coordinated and carefully timed. This paper specifically addresses this challenge, the task of generating coordinated and timed actions, as an important step towards generating videos of complex scenes. 

Current approaches for conditional video generation are not well suited to condition the generation on multiple coordinated (potentially simultaneous) timed actions. One line of work, \textit{Class Conditional Video Generation}~\citep{Tulyakov2018MoCoGAN,nawhal2020generating} relies on a single action~\citep{Tulyakov2018MoCoGAN} and a potential object class~\citep{nawhal2020generating}. However, this approach cannot deal with variable number of objects, multiple simultaneous actions and does not control the timing of the actions. \textit{Future Video Prediction}~\citep{watters2017visual,ye2019compositional}, generates future frames based on an initial input frame, but one frame will not suffice for generation of multiple coordinated actions.  
In another line of work, \textit{Video-to-Video Translation}~\citep{wang2018vid2vid,wang2018fewshotvid2vid}, the goal is to translate a sequence of semantic masks into an output video. However, segmentation maps contain only object class information, and thus do not explicitly capture the action information. As \citet{wang2018vid2vid} noted for the case of generating car turns, semantic-maps do not have enough information to distinguish car turns from a normally driving car, which leads to mistakes in video-translation. Finally, \textit{Text-to-Video}~\citep{li2018video,gupta2018imagine} work is promising because language can describe complex actions. However, in applications that require a precise description of a scene, natural language is not ideal due to ambiguities \citep{macdonald1994lexical}, user subjectivity \citep{wiebe2004learning} and difficulty to infer specific timings, if such information is missing.
Hence, we address this problem with a more structured approach. 

To generate videos based on actions, we propose to represent actions in an object-centric graph structure we call an ``Action Graph'' (AG) and define the new task of ``Action Graph to Video'' generation. An AG is a graph structure aimed at representing coordinated and timed actions. Its nodes represent objects and edges represent actions annotated with their start and end time (\Figref{fig:teaser}). AGs are an intuitive representation for describing timed actions and would be a natural way to provide precise inputs to generative models. A key advantage of AGs is their ability to describe the dynamics of objects and actions precisely in a scene. In the ``Action Graph to Video'' task, the input is an initial frame of the video with the ground-truth layout consisting of the objects classes and bounding boxes, and an AG.

There are multiple unique challenges in generating a video from an AG that cannot be addressed using current generation methods. First, each action in the graph unfolds over time, so the model needs to ``keep track'' of the progress of actions rather than just condition on previous frames as commonly done. Second, AGs may contain multiple concurrent actions and the generation process needs to combine them in a realistic way. Third, one has to design a training loss that captures the spatio-temporal video structure to ensure that the semantics of the AG is accurately captured.

To address these challenges, our AG2Vid model uses three levels of abstraction. First, an action scheduling mechanism we call ``Clocked Edges'' tracks the progress of actions over time. Based on this, a graph neural network~\citep{kipf2016semi} operates on the AG with the Clocked Edges and predicts a sequence of scene layouts. Finally, pixels are generated conditioned on the predicted layouts. We apply our AG2Vid model to CATER~\citep{girdhar2020cater} and Something-Something V2~\citep{goyal2017something} datasets and show that our approach results in videos that are more realistic and semantically compliant with the input AG than the baselines.

Last, we evaluate the AG2Vid model in a zero-shot setting where AGs involve novel compositions of the learned actions. The structured nature of our approach enables it to successfully generalize to new ``composite'' actions, as confirmed by the human raters. We also provide qualitative results of our model trained on Something-Something V2 in a challenging setting where AGs include multiple actions, while at training time each AG has a single action.

\section{Related work}
\label{sec:related_work}

Video generation is challenging  because videos often contain long range dependencies. Recent generation approaches~\citep{vondrick2016generating, denton2018stochastic,lee2018savp,babaeizadeh2018stochastic,fidelity2019villegas,Kumar2020VideoFlow} extended the framework of unconditional image generation to video, based on a latent representation. For example, MoCoGAN~\citep{Tulyakov2018MoCoGAN} disentangles the latent space representations of motion and content to generate a sequence of frames using RNNs; TGAN~\citep{TGAN2017} generates each frame in a video separately while also using a temporal generator to model dynamics across the frames. Here, we tackle a different problem by aiming to generate videos that comply with Action Graphs (AGs).

Class conditional action generation allows the generation of video based on a single action~\citep{Tulyakov2018MoCoGAN, nawhalgenerating}. A recent method, HOI-GAN (HG)~\citep{nawhalgenerating}, was proposed for the generation task of a single action and object. Specifically, HG addresses the zero-shot setting, and the model is tested on action and object compositions which are first presented at test time. Our focus is on generation of multiple simultaneous actions over time, performed by multiple objects. Our approach directly addresses this challenge via the AG representation and the notion of clocked edges.

Other conditional video generation methods have attracted considerable interest recently, with focus on two main tasks: video prediction~\citep{mathieu2015deep,battaglia2016interaction,walker2016uncertain,watters2017visual,kipf2018neural,ye2019compositional} and video-to-video translation~\citep{wang2018fewshotvid2vid,chan2019everybody,siarohin2019animating, kim2019deep, mallya2020world}. In video prediction, the goal is to generate future video frames conditioned on few initial frames. For example, it was proposed to train predictors with GANs~\citep{goodfellow2014generative} to predict future pixels~\citep{mathieu2015deep}. However, directly predicting pixels is challenging~\citep{walker2016uncertain}. Instead of pixels, researchers explored object-centric graphs and performed prediction on these~\citep{battaglia2016interaction,luc2018predinst,ye2019compositional}. While inspired by object-centric representations, our method is different from these works as our generation is goal-oriented and guided by an AG. The video-to-video translation task was proposed by~\citet{bansal2018recycle,wang2018vid2vid}, where a natural video is generated from a different video based on frame-wise semantic segmentation annotations or keypoints. However, densely labeling pixels for each frame is expensive, and might not even be necessary. Motivated by this, researchers have sought to perform generation conditioned on more accessible signals including audio or text \citep{song2018talking,fried2019text,ginosar2019learning}. Here, we propose to synthesize videos conditioned on an AG, which is cheaper to obtain compared to semantic segmentation and is a more structured representation compared to natural audio or text.

Various methods have been proposed to generate videos based on input text~\citep{marwah2017attentive,pan2017create,li2018video}. Most recent methods typically used very short captions which do not contain complex descriptions of actions. For example,~\citet{li2018video} used video-caption pairs from YouTube, where typical captions are "playing hockey" or "flying a kite". \citet{gupta2018imagine} proposed the Flinstones animated dataset and introduced the CRAFT model for text-to-video generation. While the CRAFT model relies on text-to-video retrieval, our approach works in an end-to-end manner and aims to accurately synthesize the given input actions.

 Scene Graphs (SG)~\citep{johnson2015image,johnson2018image} are a structured representation that models scenes, where objects are nodes and relations are edges. SGs have been widely used in various tasks including image retrieval~\citep{johnson2015image,schuster2015generating}, relationship modeling~\citep{referential_relationships,Schroeder2019ICCV,raboh2020dsg}, SG prediction~\citep{sg_generation_msg_pass,pixels_to_graph,zellers2018scenegraphs,herzig2018mapping}, and image captioning~\citep{xu2019scene}. SGs have also been applied to image generation~\citep{johnson2018image,deng2018probabilistic,herzig2019canonical}, where the goal is to generate a natural image corresponding to the input SG. Recently, \citet{ji2019action} presented Action Genome, a video dataset where videos of actions from Charades~\cite{sigurdsson2016hollywood} are also annotated by SGs. More generally, spatio-temporal graphs have been explored in the field of action recognition~\citep{jain2016structural,sun2018actor,Wang_videogcnECCV2018,yan2018spatial,girdhar2019video,herzig2019spatio,materzynska2019something}. For example, a space-time region graph was proposed by \citet{Wang_videogcnECCV2018} where object regions are taken as nodes and a GCN \citep{kipf2016semi} is applied to perform reasoning across objects for classifying actions. Recently, it was also shown \citep{ji2019action,yi2019clevrer,girdhar2020cater} that a key obstacle in action recognition is the ability to capture the long-range dependencies and compositionality of actions. While inspired by these approaches, we focus on generating videos which is a very different challenge.

\begin{figure*}[t!]
    \centering
    \includegraphics[width=0.9\linewidth]{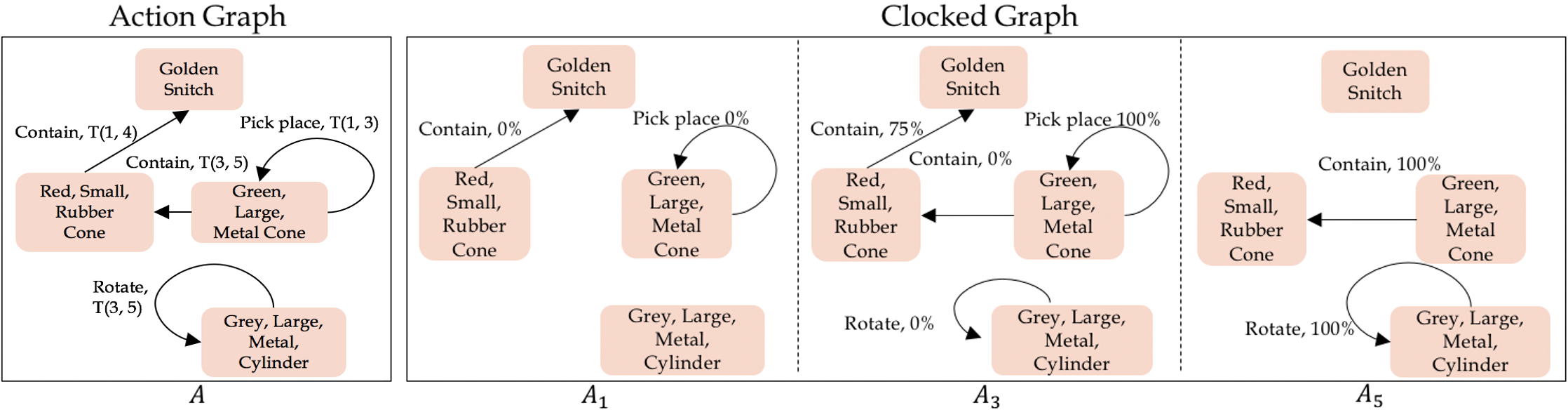}
    \vspace{-0.3cm}
    \caption{Example of a partial Action Graph execution schedule in different time-steps.}
    \label{fig:clocked_edges}
\end{figure*}



\section{Action Graphs}
\label{sec:action_graphs}
Our goal in this work is to build a model for synthesizing videos that contain a specified set of timed actions. A key component in this effort is developing a semantic representation to describe the actions performed by different objects in the scene. Towards this end, we propose to use a graph-structure we call \textit{Action Graph} (AG). In an AG, nodes correspond to objects, and edges correspond to actions that the objects participate in. Objects and actions are labeled with semantic categories, while actions are also annotated with their start and end times.  
Formally, an AG is a tuple $(\cC, \cA, O, E)$ defined as follows:
    
    \textbf{A vocabulary of object categories $\mathbf{\cC}$:} Object categories can include attributes, e.g., ``Blue Cylinder'' or ``Large Box''.
    
    \textbf{A vocabulary of actions $\mathbf{\cA}$:} Some actions, such as ``Slide'' may also have attributes (e.g., the destination coordinates).
    
    \textbf{Object nodes $\mathbf{O}$:} A set $O \in \cC^n$ of $n$ objects.  
    
    \textbf{Action edges $\mathbf{E}$:} Actions are represented as labeled directed edges between object nodes. Each edge is annotated with an action category and with the time period during which the action is performed. Formally, each edge is of the form $(i,a,j,t_s,t_e)$, where $i,j \in \{1, ... ,n\}$ are object instances, $a \in \cA$ is an action and $t_s,t_e \in \mathbb{N}$ are action start and end time. Thus, this edge implies that object $i$ (which has category $o_i$) performs an action $a$ over object $j$, and that this action takes place between times $t_s$ and $t_e$. We note that an AG edge can directly model actions over a single object and a pair of objects. For example, ``Swap the positions of objects $i$ and $j$ between time 0 and 9'' is an action over two objects corresponding to the edge $(i, swap, j, 0, 9)$. Some actions, such as ``Rotate'', involve only one object and will therefore be specified as self-loops.

\section{Action Graph to Video via Clocked Edges}
\label{sec:action_to_vid_model}

We now turn to the key challenge of this paper: transforming an AG into a video. Naturally, this transformation will be learned from data. The generation problem is defined as follows: we wish to build a generator that takes as input an AG and outputs a video. We will also allow conditioning on the first video frame and layout, so we can preserve the visual attributes of the given objects.\footnote{Using the first frame and layout can be avoided by using a SG2Image model~\citep{johnson2018image,ashual2019specifying,herzig2019canonical} for generating the first frame.}

There are multiple unique challenges in generating a video from an AG. First, each action in the graph unfolds over time, so the model needs to ``keep track'' of the progress of actions. Second, AGs may contain multiple concurrent actions and the generation process needs to combine them in a realistic way. Third, one has to design a training loss that captures the spatio-temporal video structure to ensure that the semantics of the AG is accurately captured. 

\noindent\textbf{Clocked Edges.} As discussed above, we need a mechanism  for monitoring the progress of action execution during the video generation. A natural approach is to add a ``clock'' for each action, for keeping track of action progress. See~\Figref{fig:clocked_edges} for an illustration. Formally, we keep a clocked version of the graph $\ag$ where each edge is augmented with a temporal state. Let $e=(i,a,j,t_s,t_e) \in E$ be an edge in the AG $\ag$. We define the progress of $e$ at time $t$ to be $r_t = \frac{t-t_s}{t_e - t_s}$, and clip $r_t$ to $[0,1]$
Thus, if $r_t=0$ the action has not started yet, if $r_t \in (0,1)$ it is currently being executed, and if $r_t=1$ it has completed. We then create an augmented version of the edge $e$ at time $t$, denoted $e_t=(i,a,j,t_s,t_e,r_t)$. We define $\ag_t=\{e_t | e \in \ag\}$ to be the AG at time $t$. 
To summarize the above, we take the original graph $\ag$ and turn it into a sequence of AGs $\ag_0,\ldots,\ag_T$, where $T$ is the last time-step. Each action edge in the graph now has a unique clock for its execution. This facilitates a timely execution and coordination between actions.

\begin{figure*}[t!]
    \centering
    \includegraphics[width=0.9\linewidth]{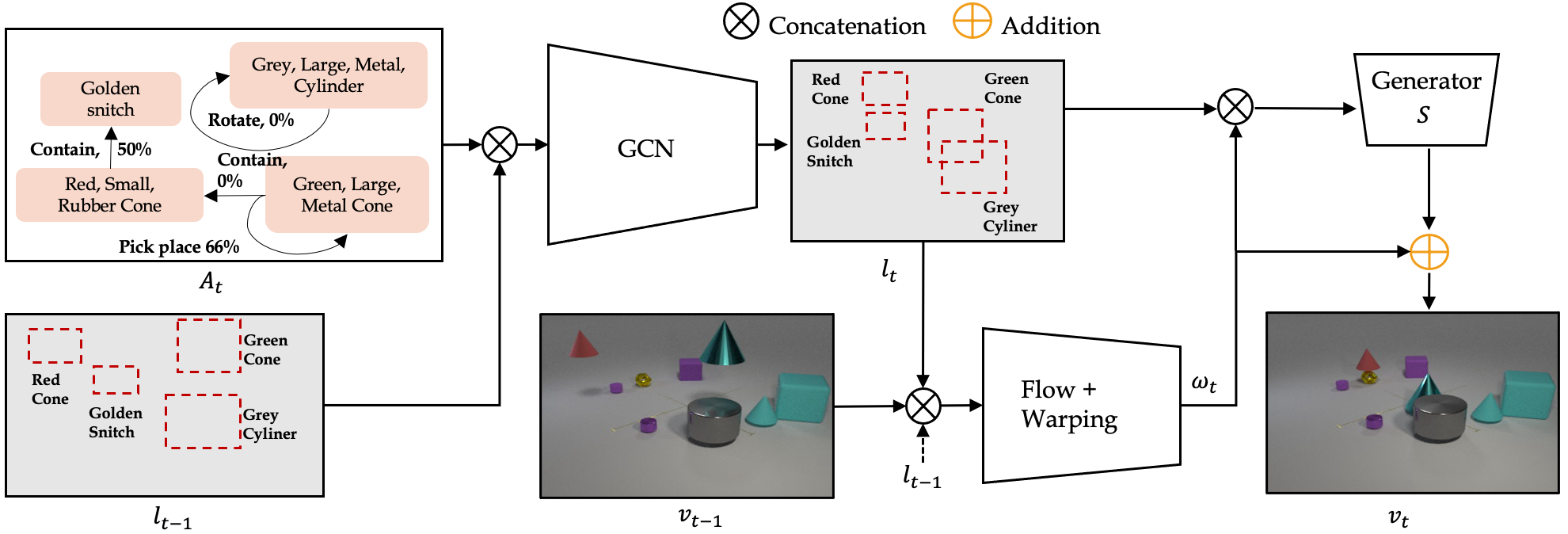}
    \vspace{-0.3cm}
    \caption{Our AG2Vid Model. The AG $\ag_t$ describes the execution stage of each action at time $t$. Together with the previous layout $\ell_{t-1}$, it is used to generate the next layout $\ell_{t}$ which has object representations that are enriched with actions information from $A_t$. Then, ${\ell}_{t}, {\ell}_{t-1}, v_{t-1}$ are used to generate the next frame.}
    \label{fig:model}
\end{figure*}

\subsection{The AG2Vid Model} 
\label{sec:model}

Next, we describe our proposed AG-to-video model (AG2Vid). \Figref{fig:model} provides a high-level illustration of the model architecture. The idea of the generation process is that the AG is first used to produce intermediate layouts, and then these layouts are used to produce frame pixels. We let $\ell_t=(x_t, y_t, w_t, h_t, z_t)$ denote the set of predicted layouts for the $n$ objects in the video at time $t$. The values $x_t,y_t,w_t, h_t \in [0,1]^{n}$ are the bounding box coordinates for all objects, and $z_t$ are the object descriptors, aimed to represent objects given the actions they participate in, as well as other actions in the scene that might affect them. 
Let $v_t$ denote the generated frame at time $t$, and $p({v_2},\ldots,{v_T}, {\ell_2}, \ldots {\ell_T}|\ag, v_1, \ell_1)$ denote the generating distribution of the frames and layouts given the AG and the first frame $v_1$ and scene layout $\ell_1$.

We assume that the generation of the frame and layout directly depends only on the current AG and the recently generated frames and layouts. Specifically, this corresponds to the following form for $p$:
\begin{align}
   p({v_2}, &..., {v_T}, {\ell_2}, ..., {\ell_T} | A, v_1, l_1) =   \\ 
    &\prod^{T}_{t=2}{p({\ell_{t}} | A_{t}, {\ell_{t-1}}) p({v_{t}} | {v_{t-1}},{\ell_t},{\ell_{t-1})}}. \nonumber
    \label{eq:prob}
\end{align}
Conditioning over $\ag_t$ provides both short and long term information which makes it possible to predict future layouts. For example, it can be inferred from $A_3$ in \Figref{fig:clocked_edges} that the ``Grey Large Metal Cylinder'' is scheduled to ``Rotate'' and thus is expected to stay in the same spatial location for the rest of the video.

We refer to the distribution $p({l_t}|\cdot)$ as \textit{The Layout Generating Function (LGF)} and to $p({v_t}|\cdot)$ as \textit{The Frame Generating Function (FGF)}. Next, we describe how we model these distributions as functions.

\noindent\textbf{The Layout Generating Function (LGF).}
At time $t$ we want to use the previous layout $\ell_{t-1}$ and current AG $\ag_t$ to predict the current layout $\ell_t$. The rationale is that $\ag_t$ captures the current state of the actions and can thus ``propagate'' $\ell_{t-1}$ to the next layout. This prediction requires integrating information from different objects
as well as the progress of the actions given by the edges of $\ag_t$. Thus, a natural architecture for this task is a \textit{Graph Convolutional Network} (GCN) that operates on the graph $\ag_t$ whose nodes are ``enriched'' with the layouts $\ell_t$. Formally, we construct a new graph of the same structure as $\ag_t$, with new features on nodes and edges. The graph nodes features are comprised of the previous object locations defined in $\ell_{t-1}$ and the embeddings of the object categories $O$.
The features on the edges are the embedding of action $a$ and the progress of the action, $r_t$, taken from the edge $(i,a,j,t_s,t_e,r_t)$ of $\ag_t$. 
Then, node and edge features are repeatedly re-estimated for $K$ steps using a GCN. The resulting activations at timestep $t$ are $z_t\in \reals^{O\times D}$ which we use as the new object descriptors. Each such object descriptor is enriched with the action information. An MLP is then applied to it to produce the new box coordinates, which together form the layout ${\ell_t}$.\footnote{For more details refer to Sec.1 in the Supplemental material.}

\noindent\textbf{The Frame Generating Function (FGF).} After obtaining the layout $\ell_t$ which contains the updated object representations $z_t$, we use it along with $v_{t-1}$ and $\ell_{t-1}$ to predict the next frame. The idea is that $\ell_t,\ell_{t-1}$ characterize how objects should move, $z_t, z_{t-1}$ capture the object features enriched with action information, and $v_{t-1}$ shows their last appearance. Combining all of them should allow us to generate the next frame accurately. As the first step, we transform the layouts $\ell_t,\ell_{t-1}$ to feature maps ${m}_{t-1},{m}_t\in \reals^{H\times W\times D}$ which have a spatial representation similar to images.
To transform a layout $l$ to a feature map $m$, we follow a similar process as in~\cite{johnson2018image}. We construct a feature map per object $\hat{m}\in \reals^{O\times H\times W\times D}$, which is initialized with zeros, and for every object assign its embedding from $z$ to its location from $l$. Finally, $m$ is obtained by summing over the objects feature maps $\hat{m}$. These feature maps provide a coarse object motion. To obtain more fine-grained motion, we estimate how pixels in the image will move from time $t-1$ to $t$ using optical flow. We compute $f_t = F(v_{t-1},m_{t-1},m_t)$, where $F$ is a flow prediction network similar to~\citep{Reda2017flownet2}. The idea is that given the previous frame and two consecutive feature maps, we should be able to predict the flow. $F$ is trained using an auxiliary loss and does not require additional supervision (see \Secref{sec:losses}). Given the flow $f_t$ and previous frame $v_{t-1}$, a natural estimate of the next frame is to use a warping function~\citep{appFlowZhou16} $w_t = W(f_t,v_{t-1})$. Finally we refine $w_t$ via a network $S(m_{t}, w_t)$ that provides an additive correction resulting in the final frame prediction: $v_{t} = w_t + S(m_{t}, w_t)$, where the $S$ network is the SPADE generator~\citep{park2019semantic}.


\begin{figure*}[t!]
    \centering
    \href{https://github.com/roeiherz/AG2Video/blob/master/Videos/Figure4.gif}{\includegraphics[trim=0 20 0 0,clip,width=0.9\linewidth]{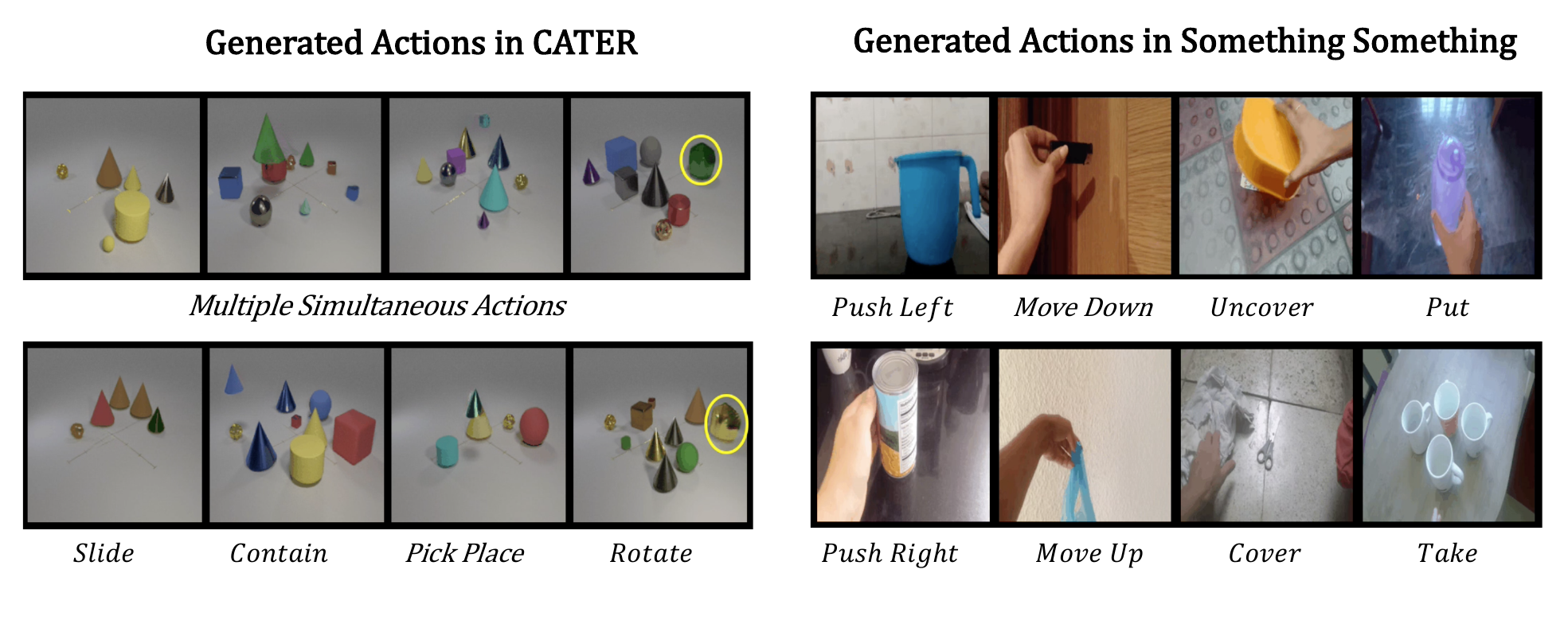}}
    \vspace{-0.5cm}
    \caption{Qualitative examples of generation on CATER and Something-Something V2. AG2Vid generated videos of four and eight standard actions on CATER and Something-Something V2, respectively. For CATER we also used AGs with multiple simultaneous actions, and the generated actions indeed correspond to those (verified manually). For more examples please refer to \Figref{fig:cater_actions} and \Figref{fig:smth_actions} in the Supplemental. \textbf{Click on the image to play the video clip in a browser}.
    }
    \label{fig:actions}
\end{figure*}

\subsection{Losses and Training}
\label{sec:losses}
We use ground truth frames $v_t^{GT}$ and layouts $\ell^{GT}_t$ for training,\footnote{With slight abuse of notation, here we ignore the latent object descriptor part of $\ell_t$.} and rely on the following losses:\newline
\noindent\textbf{Layout Prediction Loss $\mathcal{L}_{\ell}$.} This loss is defined as  $\mathcal{L}_{\ell} = \| \ell_{t} - \ell_{t}^{GT} \|_1$, an $L_1$ loss between ground-truth bounding boxes $\ell_t^{GT}$ and predicted boxes $\ell_t$.
\newline
\noindent\textbf{Pixel Action Discriminator Loss $\mathcal{L}_{A}$.} We employ a GAN loss that uses a discriminator to distinguish the generated frames $v_t$ from the GT frames $v^{GT}_t$ conditioned on $A_t$ and $\ell_t^{GT}$. The purpose of this discriminator is to ensure that each generated frame matches a real image and respective actions $\ag_t$ and layout $\ell_t^{GT}$. Let $D_A$ be a discriminator with output in $(0,1)$. First, we follow a similar process as described in the previous section, to obtain a feature map $m_t$. For this, we construct a new graph based on $\ag_t$ and $\ell_t^{GT}$, and apply a GCN to obtain object embeddings $z_t$. Then, $m_t$ is constructed using $\ell_t^{GT}$ and $z_t$.
The input frame and feature map are then channel-wise concatenated and fed into a multi-scale PatchGAN discriminator~\citep{wang2018pix2pixHD}. The loss is then the GAN loss \citep[e.g., see][]{pix2pix2017}:
\begin{align}
    \mathcal{L}_{A} =&  \max_{D_A} \mathbb{E}_{GT} \left[\log D_A(A_t, v_t^{GT}, \ell^{GT}_{t})\right] + \\ 
    & \mathbb{E}_{p} \left[\log (1 - D_A(A_t, v_t, \ell^{GT}_{t}))\right], \nonumber
\label{eq:disc_loss}
\end{align}
where $GT$/$p$ correspond to sampling frames from the real/generated videos, respectively. Optimization is done in the standard way of alternating gradient ascent on $D_A$ parameters and descent on generator parameters.
\newline
\noindent\textbf{Flow Loss $\mathcal{L}_{f}$.}  This loss measures the error between the warps of the previous frame and the ground truth of the next frame $v^{GT}_{t}$:  $\mathcal{L}_{f} = \frac{1}{T-1}\sum_{t=1}^{T-1} \big\|w_t - v_{t} \|_1$, where $w_{t} = W(f_{t}, v_{t-1})$, as defined in Section 4.1. The loss was proposed by~\citet{appFlowZhou16,wang2018vid2vid}.
\newline
\noindent\textbf{Perceptual and Feature Matching Loss $\mathcal{L}_{P}$.} We use these losses as proposed in pix2pixHD~\citep{wang2018pix2pixHD,larsen16similar} and other previous works. 

The overall optimization problem is to minimize the weighted sum of the losses:
\begin{equation}
    \min_{\theta}\max_{D_A} \mathcal{L}_{A}(D_A) + \lambda_{\ell} \mathcal{L}_{\ell} + \lambda_{f} \mathcal{L}_{f} + \lambda_{P} \mathcal{L}_{P}\,,
\end{equation}
where $\theta$ are the trainable parameters of the generative model.

\section{Experiments and Results}
\label{sec:experiments}
We evaluate our AG2Vid model on two datasets: \textit{CATER}~\citep{girdhar2020cater} and \textit{Something-Something V2}~\citep{goyal2017something} (henceforth SmthV2). For each dataset, we learn an AG2Vid model with a given set of actions. We then evaluate the visual quality of the generated videos and measure how they semantically comply with the input actions. Finally, we estimate the generalization of the AG2Vid model to novel composed actions.

\noindent\textbf{Datasets.} We use two datasets: \textbf{(1) CATER} is a synthetic video dataset originally created for action recognition and reasoning. Each video contains multiple objects performing actions. The dataset contains bounding-box annotations for all objects, as well as the list of attributes per object (color, shape, material and size), the labels of the actions and their time. Actions include: ``Rotate'', ``Contain'', ``Pick Place'' and ``Slide''. For ``Pick Place'' and ``Slide'' we include the action destination coordinates $(x,y)$ as an attribute. We use these actions to create Action Graphs for training and evaluation. We employ the standard CATER training partition (3,849 videos) and split the validation into 30\% val (495 videos) and use the rest for testing (1,156 videos).

\textbf{(2) Something-Something V2} contains real videos of humans performing basic actions. Here we use the eight most frequent actions: ``Push Left'', ``Push Right'', ``Move Down'', ``Move Up'', ``Cover'', ``Uncover'', ``Put'' and ``Take''.
Every videos has a single action and up to three different objects, coming from $122$ different object classses, including the hand which is performing the action. We use the bounding box annotations of the objects from~\citet{materzynska2019something}. We set the first frame where the objects are present as the action's start time and the last one as the end time.

\noindent\textbf{Implementation Details.} The GCN model uses $K=3$ hidden layers and an embedding layer of $128$ units for each object and action. For optimization we use ADAM~\cite{kingma2014adam} with $lr = 1e-4$ and $(\beta_1, \beta_2) = (0.5, 0.99)$. Models were trained with a batch size of $2$ which was the maximal batch size to fit on a single NVIDIA V100 GPU. For loss weights we use  $\lambda_{B} = \lambda_{F} = \lambda_{P} = 10$ and $\lambda_{A} = 1$. We use videos of 8 FPS and 6 FPS for CATER and SmthV2 and  evaluate on videos consisting of $16$ frames which correspond to spans of $2.7$ and $2$ seconds accordingly.

\begin{table}
    \centering
    \renewcommand{\tabcolsep}{1pt}

    {\small 
    \footnotesize
    \begin{tabular}{l|cc|cc}
    \hline
    \multirow{2}{*}{{AG2Vid vs. Baseline}} & \multicolumn{2}{c}{{Semantic Accuracy}} & \multicolumn{2}{c}{{Visual Quality}} \\
    \cmidrule(l{0pt}r{0pt}){2-5} 
     & {CATER} & {SmthV2} & {CATER} & {SmthV2}\\ 
    \hline
    \hline
    CVP~\citep{ye2019compositional} & $\textbf{85.7}$ & $\textbf{90.6}$ & $\textbf{76.2}$ & $\textbf{93.8}$  \\
    
    HG~\citep{nawhal2020generating} & $-$ & $\textbf{84.6}$ & $-$ & $\textbf{88.5}$ \\
    V2V~\citep{wang2018vid2vid} & $\textbf{68.8}$ & $\textbf{84.4}$ & $\textbf{68.8}$ & $\textbf{96.9}$ \\
    RNN & $\textbf{56.0}$ & $\textbf{80.6}$ & $\textbf{52.0}$ & $\textbf{77.8}$  \\
    AG2Vid-GTL & $48.6$ & $46.2$ & $42.9$ & $50.0$  \\
    \hline
    \end{tabular}
    }
    
    \caption{\textbf{Human evaluation} of action generation with respect to Semantic Accuracy and Visual Quality. Raters have to select the better image between the AG2Vid and a baseline generation method. Each number means that AG2Vid was selected as better for $X\%$ of the presented pairs. Image resolution is $256 \times 256$.} 
    
    \label{tab:results_semantic}
    \vspace{-10pt}
\end{table}

\noindent\textbf{Performance Metrics.} The AG2Vid outputs are quantitatively evaluated as follows. 
{\bf a) Visual Quality:} It is common in video generation to evaluate the visual quality of videos regardless of the semantic content. To evaluate visual quality, we use the Learned Perceptual Image Patch Similarity (LPIPS) by~\citet{zhang2018perceptual} (lower is better) over predicted and GT videos. For the SmthV2 dataset, since videos contain single actions we can report the Inception Score (IS)~\citep{Salimans2016inception} (higher is better) and Fréchet Inception Distance (FID)~\citep{Heusel2017fid} (lower is better) using a TSM~\citep{Lin_2019_ICCV} model, pretrained on SmthV2. For CATER, we avoid reporting FID and IS as these scores rely on models pre-trained for single-action recognition, and CATER videos contain multiple simultaneous actions.
Finally, to assess the visual quality, we present human annotators with our model and a baseline generated results pairs and ask them to pick the video with the better quality.
{\bf b) Semantic Accuracy}: The key goal of AG2Vid is to generate videos that accurately depict the specified actions. To evaluate this, we ask human annotators to select which of the two given video generation models provides a better depiction of actions shown in the real video. The protocol is similar to the visual quality evaluation above. We also evaluate action timing as discussed below.

\begin{table}
\centering
\renewcommand{\tabcolsep}{1pt}
    \small
    \begin{tabular}{c|cccc}
    \hline
    \multirow{2}{*}{{Loss}} & \multicolumn{1}{c|}{{Inception $\uparrow$}} & \multicolumn{1}{c|}{{FID $\downarrow$}} & \multicolumn{2}{c}{{LPIPS $\downarrow$}}\\
    & {SmthV2} & {SmthV2} & {CATER} & {SmthV2}\\ 
    \hline
    \hline

    Flow & {1.59 $\pm$ 0.02} & {107.26 $\pm$ 1.46} & {0.14 $\pm$ 0.01} & {0.70 $\pm$ 0.06} \\
    +Percept. & {2.21 $\pm$ 0.07} & {71.70 $\pm$ 1.46} & {0.08 $\pm$ 0.03} & {0.29 $\pm$ 0.07} \\
    +Act Disc. & \textbf{3.02 $\pm$ 0.11} & \textbf{66.70 $\pm$ 1.29} & \textbf{0.07 $\pm$ 0.02} & \textbf{0.25 $\pm$ 0.08} \\
    
    \hline
    \end{tabular}
    \vspace{-3pt}
    \caption{Ablation experiment for losses of the frame generation. Losses are added one by one.}
    \vspace{-10pt}
    \label{tab:ablations}
\end{table}

\noindent\textbf{Compared Methods.} Generating videos based on AG input is a new task. There are no off-the-shelf models that can be used for direct comparison with our approach, since no existing models take AG as input and output a video. To provide fair evaluation, we compare with two types of baselines. (1) Existing baseline models that share some functionality with AG2Vid. (2) Variants of the AG2Vid model that shed light on its design choices. Each baseline serves to evaluate specific aspects of the model, as described next. \textbf{Baselines:} 
\textbf{(1) HOI-GAN (HG)~\citep{nawhalgenerating}} generates videos given a \emph{single} action-object pair, an initial frame and a layout. It can be viewed as operating on a two-node AG without timing information. We compare HG to AG2Vid on the SmthV2 dataset because it contains exactly such action graphs. HG is not applicable to CATER since the videos there contain multiple action-object pairs. \textbf{(2) CVP~\citep{ye2019compositional}} uses the first image and layout as input for future frame prediction without access to action information. CVP allows us to asses the visual quality of the AG2Vid videos. However, it is not expected that CVP captures the semantics of the action-graph, unless the first frame and action are highly correlated (e.g., a hand at the top-left corner always moves down). \textbf{(3) V2V~{\citep{wang2018vid2vid}}}: This baseline uses a state-of-the-art Vid2Vid model to generate videos from {\em ground-truth} layouts. Since it uses ground-truth layouts it provides an upper bound on Vid2Vid performance for this task. We note that Vid2Vid cannot use the AG, and thus it is not provided as input. 
\textbf{AG2Vid variants:} \textbf{(4) RNN}: This AG2Vid variant replaces the LGF GCN with an RNN that processes the AGs edges sequentially. The frame generation part is the same as in AG2Vid. The motivation behind this baseline is to compare the design choice of the GCN to a model that processes edges sequentially (RNN). \textbf{(5) AG2Vid-GTL:} An AG2Vid model that uses GT layouts at inference time. It allows us to see if using the GT layouts for all frames improves an overall AG2Vid video quality and semantics.

\begin{table}[t!]
    \centering
    {
    \resizebox{\linewidth}{!}{
    \begin{tabular}{l|cccc}
    \hline
    \multirow{2}{*}{{Methods}} &
    \multicolumn{1}{c}{{Inception $\uparrow$}} & 
    \multicolumn{1}{|c}{{FID $\downarrow$}} &
    \multicolumn{2}{|c}{{LPIPS $\downarrow$}} \\
    \cmidrule(r{0pt}){2-5}
    & {SmthV2} & {SmthV2} & {CATER} & {SmthV2} \\
    \hline
    \hline
    
    \multicolumn{5}{c}{64x64} \\
    \hline
    Real Videos & {3.90 $\pm$ 0.120} & {0.00 $\pm$ 0.00} & {0.00 $\pm$ 0.00} & {0.00 $\pm$ 0.00} \\ %
    \cline{1-5}
    HG~\citep{nawhal2020generating} & {1.66 $\pm$ 0.03} & {35.18 $\pm$ 3.60} & { - } & {0.33 $\pm$ 0.08} \\
    \textbf{AG2Vid} (Ours) & \textbf{2.51 $\pm$ 0.08} & \textbf{26.05 $\pm$ 0.73} & \textbf{0.04 $\pm$ 0.01} & \textbf{0.13 $\pm$ 0.01} \\
    \hline
    \hline

    \multicolumn{5}{c}{256x256} \\
    \hline
    Real Videos & {7.58 $\pm$ 0.20} & {0.00 $\pm$ 0.00} & {0.00 $\pm$ 0.00} & {0.00 $\pm$ 0.00} \\ 
    \cline{1-5}
    CVP~\citep{ye2019compositional} & {1.92 $\pm$ 0.03} & {67.77 $\pm$ 1.43} & {0.24 $\pm$ 0.04} & {0.55 $\pm$ 0.08} \\
    RNN & {1.99 $\pm$ 0.05} & {74.17 $\pm$ 1.54} &  {0.14 $\pm$ 0.05} & {0.26 $\pm$ 0.08} \\
    V2V~\citep{wang2018vid2vid} & {2.22 $\pm$ 0.07} & {67.51 $\pm$ 1.42} & {0.11 $\pm$ 0.02} & {0.29 $\pm$ 0.09} \\
    \cline{1-5}
    \textbf{AG2Vid} (Ours) & \textbf{3.02 $\pm$ 0.11} & \textbf{66.70 $\pm$ 1.29} & \textbf{0.07 $\pm$ 0.02} & \textbf{0.25 $\pm$ 0.08} \\
    \textbf{AG2Vid-GTL} (Ours) & \textbf{3.52 $\pm$ 0.14} & \textbf{65.04 $\pm$ 1.25} & \textbf{0.06 $\pm$ 0.02} & \textbf{0.22 $\pm$ 0.09} \\
    \hline
    \end{tabular}
    }
    }
    \vspace{-7pt}
    \caption{Visual quality metrics of conditional video-generation methods in CATER and \textit{SmthV2}. All methods use resolution $256 \times 256$ except for HG, which only supports $64 \times 64$.} 
    \label{tab:results_visual} 
    \vspace{-12pt}
\end{table}

\begin{figure}[t!]
      \centering
      \href{https://github.com/roeiherz/AG2Video/blob/master/Videos/Figure5.gif}{\includegraphics[width=\linewidth]{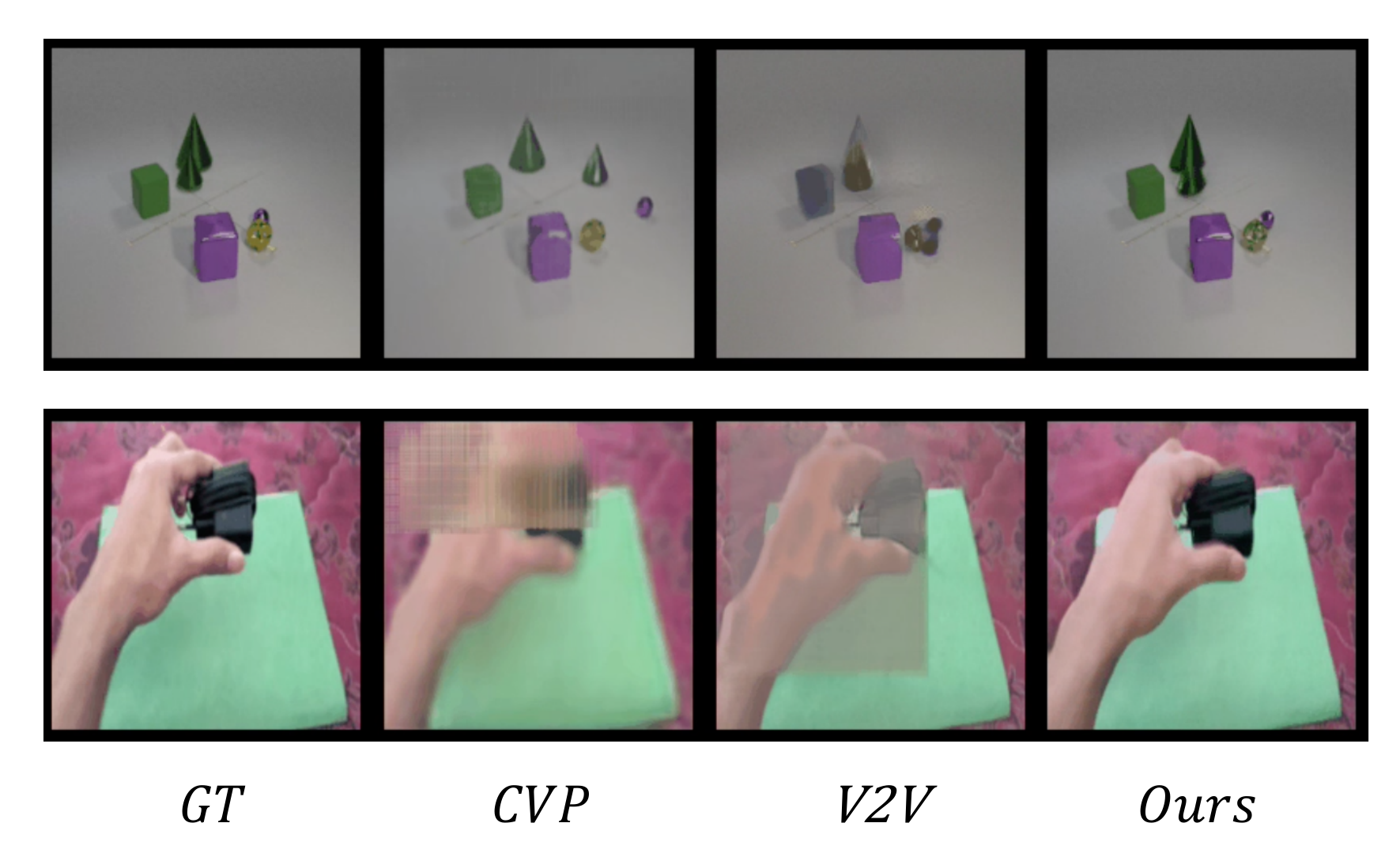}}
      \caption{Comparison to the baselines methods. The top row is from \textit{CATER}, the bottom row is from \textit{Something-Something V2}. \textbf{Click on the image to play the video clip in a browser}.}
      \label{fig:results}
    \vspace{-18pt}
\end{figure}

\begin{figure}[t!]
      \centering
      \href{https://github.com/roeiherz/AG2Video/blob/master/Videos/Figure6.gif}{\includegraphics[width=\linewidth]{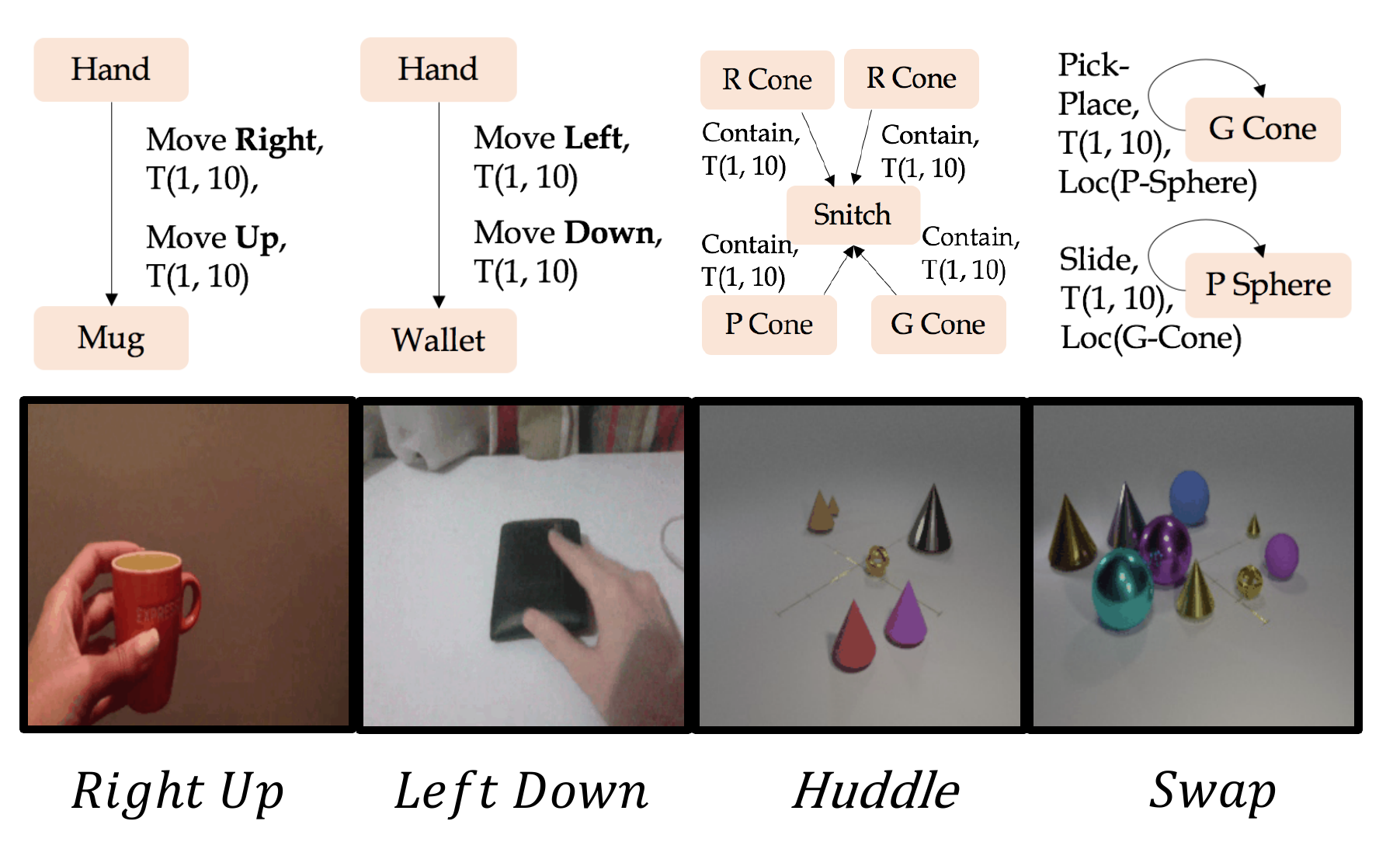}}
      \vspace{-10pt}
      \captionof{figure}{Zero-shot ``composite'' actions in \textit{Something-Something V2} and \textit{CATER}. For example, the ``Swap'' action is composed by performing the ``Pick-Place'' and ``Slide'' actions simultaneously. \textbf{Click on the image to play the video clip in a browser}.
      \vspace{-5pt}
    }
      \label{fig:compositional_smth_cater}
\end{figure}

{\bf Layout Generation Ablation Results.} We compare the choice of GCN to the RNN baseline, as an alternative to the implementation of the AG2Vid LGF. To evaluate, we use the mIOU (mean Intersection Over Union) between the predicted and GT bounding boxes. The results confirm the advantage of GCN over RNN with a mIOU of $93.09$ vs. $75.71$ over CATER and $51.32$ vs. $41.28$ over SmthV2. We include the full results in Sec.~\ref{sec:rnn_baseline} in the Supplementary.
\newline
{\bf Loss Ablation Results.} \tabref{tab:ablations} reports ablations over the losses used in FGF, confirming that each loss, including the pixel actions discriminator loss, contributes to the overall visual quality on CATER and SmthV2.
\newline
{\bf Semantic and Visual Quality.} 
We include human evaluation results in \tabref{tab:results_semantic}, which indicate that AG2Vid is more semantically accurate and has better visual quality than the baselines. Comparing to AG2Vid-GTL, it is slightly worse, as expected. We also provide qualitative AG2Vid generation examples in \Figref{fig:actions}, and comparison to baselines in \Figref{fig:results}. For more results and a per-action analysis of the correctness of the generated actions see the Suppl.\Secref{sec:semantic_eval}. \tabref{tab:results_visual} evaluates visual quality using several automatic metrics, with similar takeaways as in \tabref{tab:results_semantic}.
\newline
{\bf Action Timings.} To evaluate to what extent the AG2Vid model can control the timing of actions, we generated pairs of AGs where in the first AG an action is coordinated to appear earlier than in the second AG, keeping everything else constant. We then use the AG2Vid model to create corresponding videos and ask MTurk annotators to choose the video where the action appears first. In 89.45\% of the cases, the annotators confirm the intended result. For more information see \Secref{sec:timing} in the Supplementary. 
\newline
{\bf Composing New Actions.} Finally, we validate the ability of our AG2Vid model to generalize at test time to zero-shot compositions of actions. Here, we manually define four new ``composite'' actions. We are using learned atomic actions to generate new combinations that did not appear in the training data (either by having the same object perform multiple actions at the same time, or having multiple objects perform coordinated actions), see \Figref{fig:compositional_smth_cater}. Specifically, in CATER, we created the action ``Swap'' based on ``Pick-Place" and ``Slide", and action ``Huddle" based on multiple ``Contain" actions. In SmthV2 we composed the ``Push-Left'' and ``Move-Down'' to form the ``Left-Down'' action, and ``Push-Right'' with ``Move-Up'' to form the ``Right-Up'' action. For each generated video, raters were asked to choose the correct action class from the list of all possible actions.
The average class recall for CATER and SmthV2 is $96.65$ and $87.5$ respectively. See Supplementary for more details.

\begin{figure}[t!]
      \centering
      \href{https://github.com/roeiherz/AG2Video/blob/master/Videos/Figure7.gif}{\includegraphics[width=\linewidth]{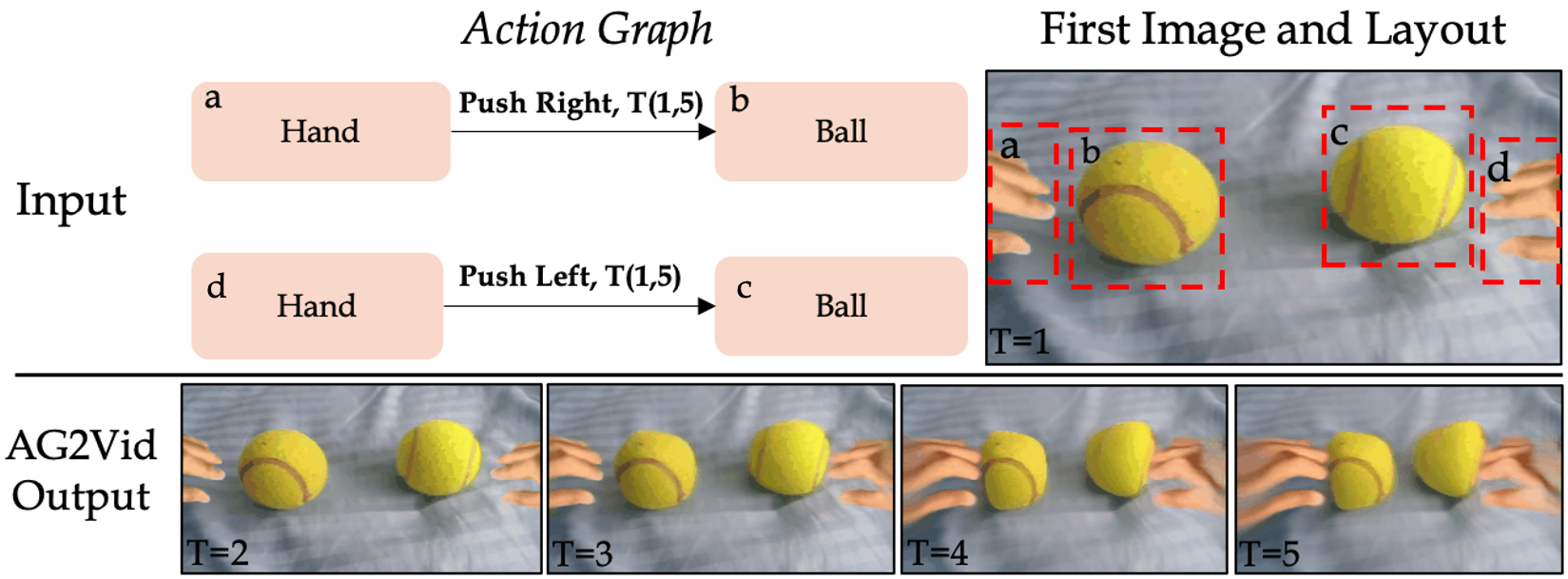}}
      \vspace{-10pt}
      \captionof{figure}{AG2Vid generation example of a video with four objects and two simultaneous actions. The hands (a,d) were not part of the original image and were manually added. Here, we illustrate the ability of our model to work on AGs that are much more complex than the typical \textit{Something-Something V2} examples.}
      \vspace{-5pt}
      \label{fig:in_the_wild}
\end{figure}

\begin{figure}[t!]
      \centering
      \href{https://github.com/roeiherz/AG2Video/blob/master/Videos/Figure8.gif}{\includegraphics[width=\linewidth]{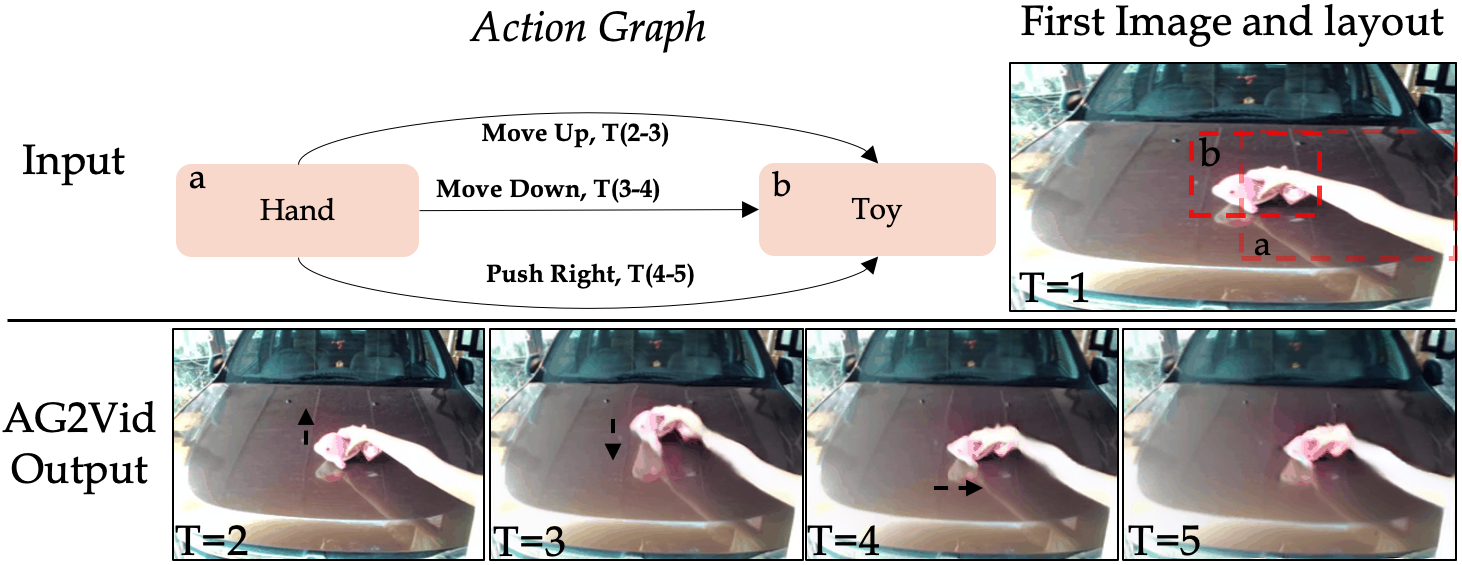}}
      \vspace{-10pt}
      \captionof{figure}{AG2Vid generation example of a video with three consecutive actions. Here, we illustrate the ability of our model to work on AGs with multiple actions, while trained on single actions. \textbf{Click on the image to play the video clips in a browser}.}
      \vspace{-5pt}
      \label{fig:in_the_wild2}
\end{figure}

To further demonstrate that AG2Vid can generalize to AGs with more objects and actions than seen in training, we include additional examples. In \Figref{fig:in_the_wild}, the AG contains two simultaneous actions that were never observed together during the training. In \Figref{fig:in_the_wild2}, the AG contains three consecutive actions, a scenario also not seen during training.
These examples demonstrate the ability of our AG2Vid model to generalize to new and more complex scenarios.

\section{Discussion}
\label{sec:discussion}
In this work, we aim to generate videos of multiple coordinated and timed actions. To represent the actions, we use an Action Graph, and propose the ``Action Graph to Video'' task. Based on the AG representation, our AG2Vid model can synthesize complex videos of new action compositions and control the timing of actions. The AG2Vid model uses three levels of abstraction: scheduling actions using the ``Clocked Edges'' mechanism, coarse layout prediction, and fine-grained pixels level prediction. Despite outperforming other methods, AG2Vid still fails in certain cases, e.g, when synthesize occluded or overlapping objects. We believe more fine-grained intermediate representations like segmentation masks can be incorporated to alleviate this difficulty. Furthermore, while it was shown that AG2Vid can synthesize actions that have rigid motion, synthesizing actions with non-rigid motion might require conditioning on random noise, or by explicitly defining action attributes like speed or style. We leave such extensions for future work.

\section*{Acknowledgements}
We would like to thank Lior Bracha for her help running MTurk experiments, and to Haggai Maron and Yuval Atzmon for helpful feedback and discussions. This project has received funding from the European Research Council (ERC) under the European Unions Horizon 2020 research and innovation programme (grant ERC HOLI 819080). Trevor Darrell’s group was supported in part by DoD including DARPA's XAI, LwLL, and/or SemaFor programs, as well as BAIR's industrial alliance programs. Gal Chechik's group was supported by the Israel Science Foundation (ISF 737/2018), and by an equipment grant to Gal Chechik and Bar-Ilan University from the Israel Science Foundation (ISF 2332/18). This work was completed in partial fulfillment for the Ph.D degree of Amir Bar.

\bibliography{egbib}
\bibliographystyle{icml2021}

\newpage
\setcounter{section}{0} 

\title{Supplementary Material for ``Compositional Video Synthesis with Action Graphs''}

\maketitle
In this supplementary file we provide additional information about our model, training losses, experimental results, and qualitative examples. Specifically, Supplementary Sections~\ref{sec:gcn} and~\ref{supp:loss} provide additional information about the AG2Vid model GCN implementation and losses, and in Section~\ref{supp:actions} we provide more AG2Vid action generation results. Finally, in Section~\ref{supp:experiments} we provide additional baseline implementation details and more evidence regarding AG2Vid performance.

\section{Graph Convolution Network}
\label{sec:gcn}
As explained in the main paper, we used a Graph Convolution Network (GCN)~\citep{kipf2016semi} to predict the layout $\ell_{t}$ at time step $t$. The GCN uses the structure of the action graph, and propagates information along this graph (in $K$ iterations) to obtain a set of layout coordinates and embedding per object.

Each object category $c\in\cC$ is assigned a learned embedding $\phiv_c\in\reals^D$ and each action $a\in\cR$ is assigned a learned embedding $\psiv_a\in\reals^D$. We next explain how to obtain the layouts $\ell_t$ using a GCN. Consider the action graph $A_t$ at time $t$ with the corresponding clocked edges $(i,a,j,t_s, t_e, r_t)$. Denote the layout for node $i$ at time $t$ by $\ell_{t,i}$. The GCN iteratively calculates a representation for each object and each action in the graph. Let $\zz^{k}_{t,i}\in\reals^D$ 
be the representation of the $i^{th}$ object in the $k^{th}$ layer of the GCN at time $t$. Similarly, for each edge in $A_t$ given by $e=(i,a,j,t_s, t_e, r_t)$ let $\uu^{k}_{t,e}\in\reals^D$ be the representation of this edge in the $k^{th}$ layer at time $t$. These representations are calculated as follows. At the GCN input, we set the representation for node $i$ to be: $\zz^{0}_{t,i} = [\phiv_{o(i)},\ell_{t-1,i}]$. And, for each edge $e=(i,a,j,t_s, t_e, r_t)$ set $\uu^{0}_{t,e} = [\psiv_{a},r_t,\ell_{t-1,i},\ell_{t-1,j}]$.

All representations at time $0$ are transformed to $D$ dimensional vectors using an MLP. Next, we use three functions (MLPs) $F_s,F_a,F_o$, each from $\reals^D\times\reals^D\times\reals^D$ to $\reals^D$. These receive the subject, action and object representations and return three new representations. The updated $i^{th}$ object representation is then:\footnote{Note that a node can appear both as a ``subject'' and an ``object'' thus two different sums in the denominator.}
\begin{align}
    \zz^{k+1}_{t, i} = & \sum_{e=(i,a,j,t_s,t_e,r_t)} F_s(\zz^{k}_{t,i},\uu^{k}_{t,e},\zz^{k}_{t, j})\quad+ \\ & \sum_{e=(j,a,i,t_s,t_e,r_t)} F_o(\zz^{k}_{t, j},\uu^{k}_{t,e},\zz^{k}_{t, i})   \nonumber
\label{eq:gcn}
\end{align}
Similarly, the representation for edge $e$ is updated via: $\uu^{k+1}_{t,e} = F_a(\zz^{k}_{t, i},\uu^{k}_{t,e},\zz^{k}_{t,j})$.

Finally, we transform the GCN representations above at each time-step $t$ to a layout $\ell_t$ as follows. Let $K$ denote the number of GCN updates. The layout coordinates of $\ell_{t,i}$ are the output of an MLP applied to $\zz^{K}_{t,i}$, which are simply the set of the predicted normalized bounding box coordinates. The object descriptor of the the $i^{th}$ object is $\zz_{t,i}^K$ and we denote the final object descriptors matrix at time $t$ as $\zz_t\in \reals^{O\times D}$. The predicted bounding box coordinates are trained via the $L_1$ loss as described in Section 2 of the main paper.

\begin{figure*}[t!]
    \centering
    \href{https://github.com/roeiherz/AG2Video/blob/master/Videos/Supp_Figure1.gif}{\includegraphics[width=\linewidth]{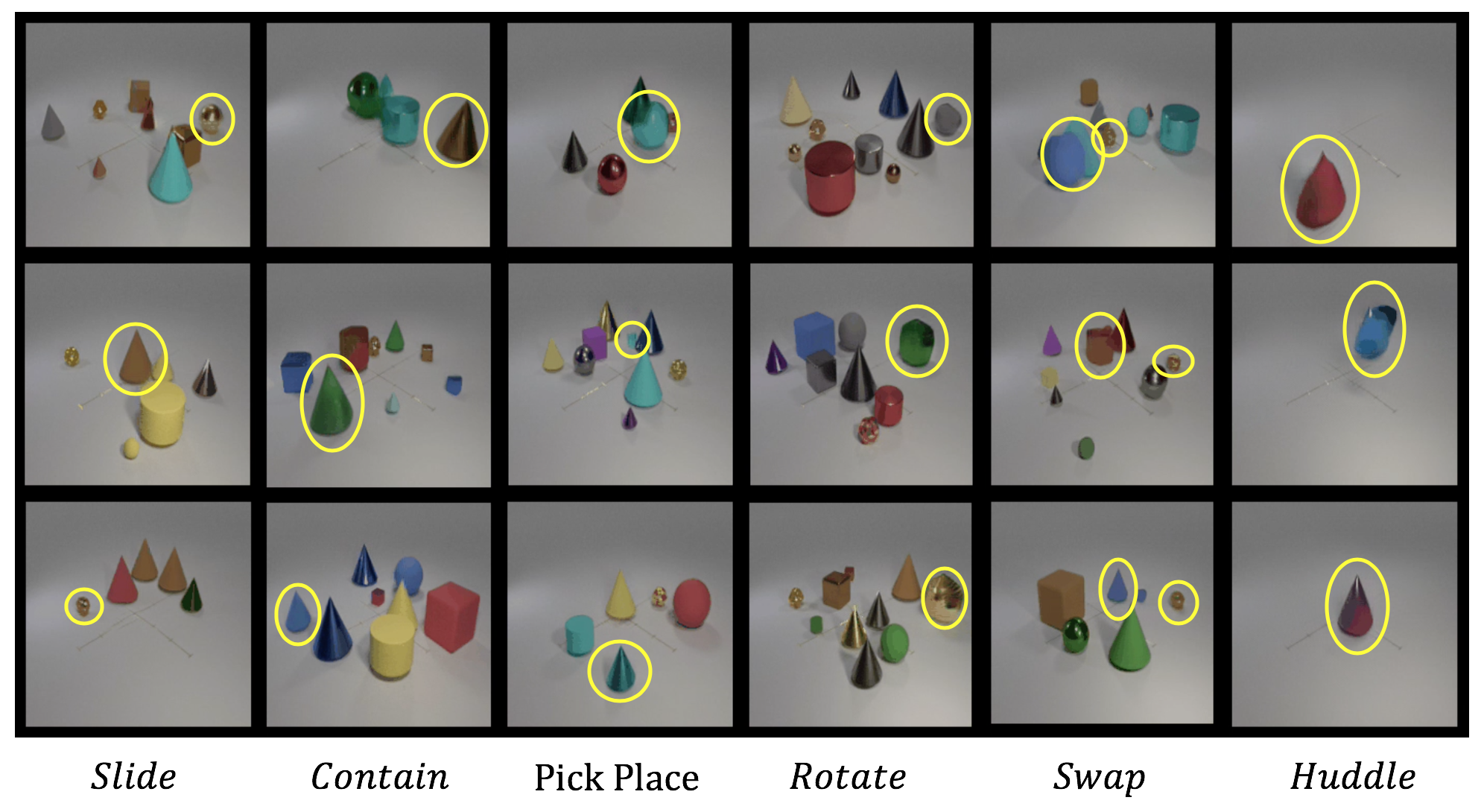}}
    \caption{Qualitative examples for the generation of actions on the CATER dataset. We use the AG2Vid model to generate videos of four standard actions and two composed unseen actions (``Swap'' and ``Huddle''). The objects involved in actions are highlighted. \textbf{Click on the image to play the video clip in a browser}.}
    \label{fig:cater_actions}
\end{figure*}


\section{Losses and Training}
\label{supp:loss}
We elaborate on the Perceptual loss and Feature Matching loss from Section 4.2.




\paragraph{Perceptual and Feature Matching loss $\mathcal{L}_{P}$.} 
We use these losses as proposed in pix2pixHD~\citep{wang2018pix2pixHD}. In particular, we use the VGG network~\citep{simonyan2014vgg} and our action discriminator $D_A$ network as feature extractors and extract features from $L$ layers given generated and ground truth images. We then minimize the error between the extracted features generated and ground truth features.

\begin{equation}
    \mathcal{L}_{P} = \sum_{l}^{L}\frac{1}{P_l}||\phi^{(l)}(v_{t})-\phi^{(l)}(v^{GT}_{t})||_1\,,
\label{eq:vgg_loss}
\end{equation}
where $\phi^{(l)}$ denotes the $l$-th layer with $P_l$ elements of the VGG/Discriminator networks. We sum the above over all frames in the videos.

The overall optimization problem is to minimize the weighted sum of the losses:
\begin{equation}
    \min_{\theta}\max_{D_A} \mathcal{L}_{A}(D_A) + \lambda_{\ell} \mathcal{L}_{\ell} + \lambda_{f} \mathcal{L}_{f} + \lambda_{P} \mathcal{L}_{P}\,,
\label{eq:loss}
\end{equation}
where $\theta$ are all the trainable parameters of the generative model, $\mathcal{L}_{\ell}$ is the Layout loss, and $\mathcal{L}_{A}$ is the pixel action discriminator loss from Section 4.2.

\section{Actions}
\label{supp:actions}
\begin{figure*}[t!]
    \centering
    \href{https://github.com/roeiherz/AG2Video/blob/master/Videos/Supp_Figure2_no_repeat.gif}{\includegraphics[width=\linewidth]{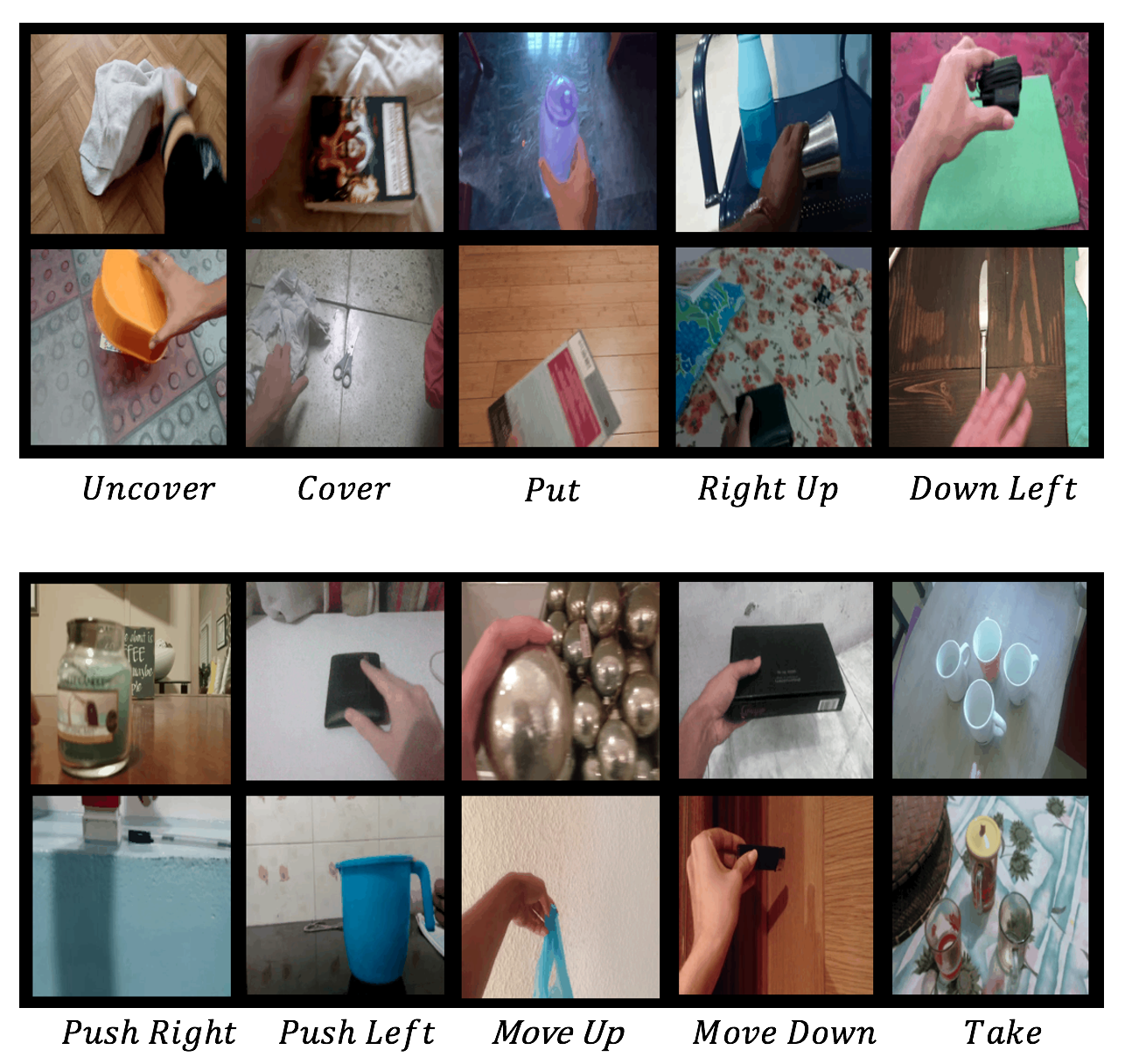}}
    \caption{Qualitative examples for the generation of actions on the Something Something dataset. We use our AG2Vid model to generate videos of eight standard actions and two composed unseen actions (``Right Up'' and ``Down Left''). \textbf{Click on the image to play the video clip in a browser}.}
    \label{fig:smth_actions}
\end{figure*}

For the \textit{Something Something} dataset~\citep{goyal2017something}, we use the eight most frequent actions. These include: ``Putting [something] on a surface'', ``Moving [something] up'', ``Pushing [something] from left to right'', ``Moving [something] down'', ``Pushing [something] from right to left'', ``Covering [something] with [something]'', ``Uncovering [something]'', ``Taking [one of many similar things on the table]''. See Figure~\ref{fig:smth_actions} for qualitative examples. The box annotations of the objects from the videos are taken from~\cite{materzynska2019something}. 
For the CATER dataset we include the four actions as provided by \cite{girdhar2020cater}. These include: ``Rotate'', ``Cover'', ``Pick Place'' and ``Slide''. See Figure~\ref{fig:cater_actions} for qualitative examples. For ``Pick Place'' and ``Slide'' which are otherwise random, we include the action destination coordinates.


\section{Experiments and Results}
\label{supp:experiments}

\subsection{LGF rule based baseline} We implement a rule based linear trajectory model. For each SmthV2 action, we encoded a constant motion pattern, conditioned on the time. For example, the action ``Move Right'' is represented by:\\
\begin{minipage}{.5\linewidth}
\centering
\begin{equation}\centering
\nonumber
  x_{t+1} = x_t + \alpha
\end{equation}
\end{minipage}%
\begin{minipage}{.5\linewidth}
\centering
\begin{equation}
\nonumber
y_{t+1} = y_t
\end{equation}
\end{minipage}
\\
\begin{minipage}{.5\linewidth}
\centering
\begin{equation}
\nonumber
w_{t+1} = w_t
\end{equation}
\end{minipage}%
\begin{minipage}{.5\linewidth}
\centering
\begin{equation}
\nonumber
h_{t+1} = h_t
\end{equation}
\end{minipage}

Where $x_t,y_t,w_t,h_t$ are the object box coordinates in time $t$, which are given in $t=0$ by the initial layout coordinates. This function is applied when the action ``Move right'' is operating on the object. We used $\alpha=0.1$.

\subsection{RNN baseline}
\label{sec:rnn_baseline}
We experiment with an RNN architecture as an alternative to the GCN implementation of the LGF. This RNN has access to the same input and supervision to the GCN, namely, to $l_{t-1}$ and $A_t$. We next explain how to obtain the layouts $\ell_t$ using the RNN. 

We follow the same notations as in Suppl Section~\ref{sec:gcn}. To obtain the $m^{th}$ object representation at time $t$, every edge $e=(i,a,j,t_s,t_e,r_t)$ and the $m^{th}$ object are mapped to a vector of features $(\zz^{0}_{t,m}, \zz^{0}_{t,i}, \uu^{0}_{t,e}, \zz^{0}_{t,j}) \in \reals^{4D}$, where the initial step $\zz$'s and $\uu$ are initialized as described in Section~\ref{sec:gcn}. We denote $U_m\in \reals^{E\times4D}$ as the matrix of edge features, where each row corresponds to the features of an edge and the $m^{th}$ object features. Next, to learn updated object representations $\zz_{t,m}$, we apply $K$ RNN layers over the sequence of edges $U_m$ and denote $\zz^{K}_{t,m}$ as the last hidden state of the $K^{th}$ RNN layer. To obtain bounding box coordinates, we then apply an MLP over $\zz^{K}_{t,m}$, similarly as in the Suppl Section~\ref{sec:gcn}. Next, we discuss the layout accuracy results, which are included in~\tabref{tab:lgf_ablations}.

\subsection{Layout Accuracy}
AG2Vid produces bounding boxes of objects as a function of time. Since our datasets contain ground-truth boxes, we can compare our predictions to these. We evaluate the intersection over union (IOU) over the predicted and ground truth boxes. We report the mean intersection over union (mIOU) which is the mean over the entire set of boxes. Additionally, we measure the recall and consider a prediction to be a ``hit'' if the IOU between the GT and predicted box is higher than 0.3 (R@0.3) or 0.5 (R@0.5). Results are reported in \tabref{tab:lgf_ablations}. While better than random, the Rule Based baseline is worse than GCN and RNN, mainly because it doesn't take into account factors like the initial relative location, the hand position, and the object size. The RNN baseline underperforms compared to the GCN, supporting the choice of GCN for layout generation in the AG2Vid model. The RNN is likely to under-perform as it assumes order over the AG list of edges, which is not a natural way to process a graph.

\begin{table}
    \tiny
  \centering
  \begin{adjustbox}{width=0.8\linewidth}
    
    \begin{tabular}{c|c|c|c}
            \hline
            Model & mIOU $\uparrow$ & R@0.3 $\uparrow$ & R@0.5 $\uparrow$ \\
            
            \hline
            \hline
            Random & 13.6 & 16.5 & 4.8 \\
            Rule Based  & 27.3 & 18.4 & 18.7  \\ 
            RNN & 41.3 & 61.7 & 39.2 \\ 
            GCN & \textbf{51.3} &  \textbf{74.5} & \textbf{53.9} \\
            \hline
  \end{tabular}%
  \end{adjustbox}
  \caption{Layout generation evaluation.}
    \label{tab:lgf_ablations}
\end{table}

\subsection{Human Evaluation of Action Timing in Generated Videos}
\label{sec:timing}
\begin{figure*}[ht!]
    \centering
    \href{https://github.com/roeiherz/AG2Video/blob/master/Videos/Supp_Figure3.gif}{\includegraphics[width=\linewidth]{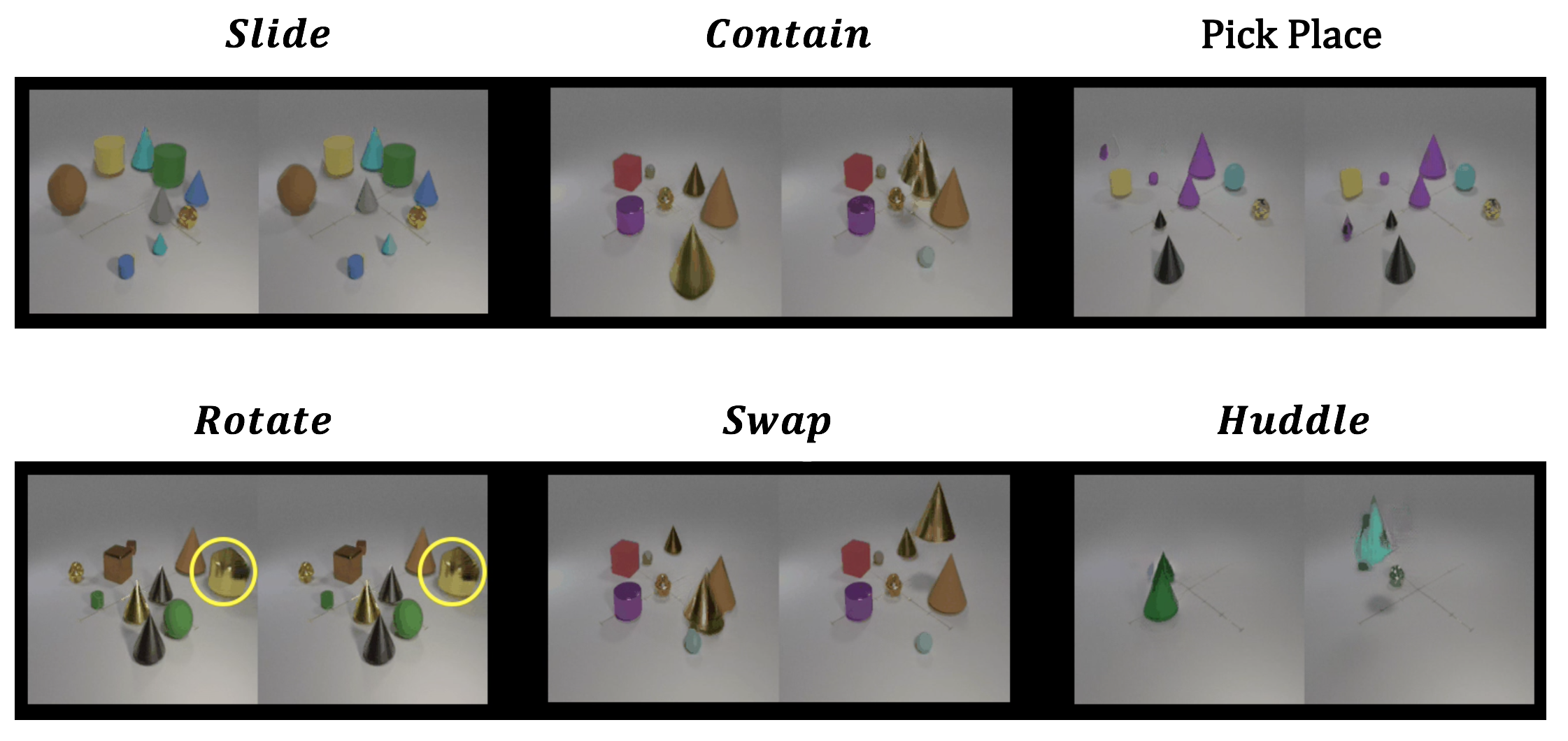}}
    \caption{Timing experiment examples in CATER. We show the clock edges can manipulate the timing of a video by controlling when the action is performed to achieve goal-oriented video synthesis. The objects involved in ``rotate'' are highlighted.  \textbf{Click on the image to play the video clip in a browser}.}
    \label{fig:cater_timing}
\end{figure*}

As described in Section 5.1, we evaluated to which extent the action graphs (AGs) can control the timing execution of actions on the CATER dataset. Thus, we generated 90 pairs of action graphs where the only difference between the two graphs is the timing of one action. We then asked the annotators to select the video where the action is executed first. The full results are depicted in Table~\ref{tab:timing_table}, and visual examples are shown in Figure~\ref{fig:cater_timing}. The results for all actions but ``Rotate'' are consistent with the expected behavior, indicating that the model correctly executes actions in a timely fashion. The ``Rotate'' action is especially challenging to generate, perhaps because it involves a relatively subtle change in the video.

\begin{table}
\small
   \centering
    \begin{tabular}{cccc}
    \hline
    \multicolumn{4}{c}{{Composed Actions}} \\

    Swap & Huddle & RU & DL\\ 
    \hline \hline

    92.1 & 98.6 & 75.0 & 100. \\
    \hline 

    \end{tabular}
    \caption{Human evaluation of the \textbf{semantic accuracy} of the actions in the generated videos.}
\label{tab:composing} 
\end{table}

\begin{table}
   \centering
   \begin{adjustbox}{width=\columnwidth}

    \begin{tabular}{cccc|cc}
    \hline
    \multicolumn{4}{c}{{Standard Actions}} & \multicolumn{2}{|c}{{Composed Actions}} \\

    Slide & Contain & Pick Place & Rotate & Swap & Huddle\\ 
    \hline \hline

    
    96.7 & 100.0 & 90.0 & 56.7 & 93.3 & 100.0 \\
    \hline 

    \end{tabular}
    
    \end{adjustbox}
    \caption{Human evaluation of \textbf{timing} in generated videos (see \secref{sec:timing}). The table reports accuracy of human annotator answer with respect to the true answer. The average accuracy over all actions is $89.45$.}
    \label{tab:timing_table} 
\end{table}

\subsection{Human Evaluation of the Semantic Quality in Generated Videos of Single Actions}
\label{sec:semantic_eval}
We provide additional evidence regarding the AG2Vid model performance in a more controlled setting, given videos that contain only single actions. To do this, we generated twenty videos per action for the SmthV2 dataset and asked three different human annotators to evaluate each video. Each annotator was asked to pick the action that best describes the video out of the list of possible actions. We provide the results in~\tabref{tab:semantic_table}. Each cell in the table corresponds to the class recall of a specific action. To determine if a video correctly matches its corresponding action, we used the majority voting over the answers of all annotators.

\begin{figure}[t]
    \centering
    \href{https://github.com/roeiherz/AG2Video/blob/master/Videos/Supp_Figure6.gif}{\includegraphics[width=\linewidth]{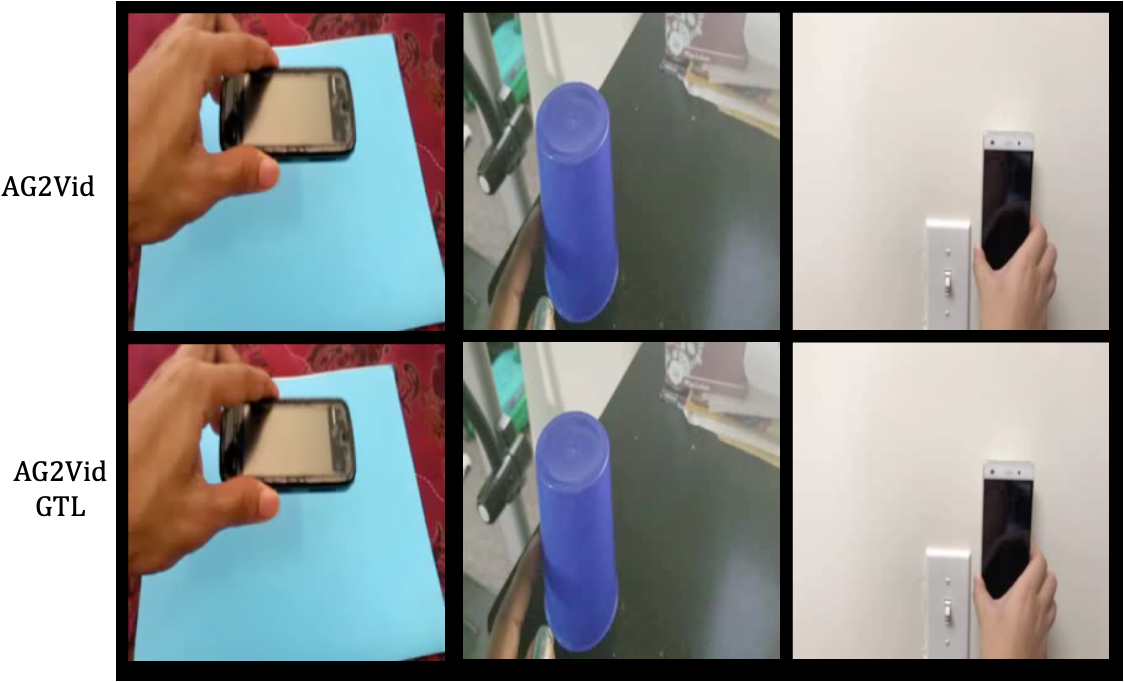}}
    \caption{Comparing AG2Vid to AG2Vid-GTL, e.g when conditioning on ground truth layout, over Something-Something V2 dataset. \textbf{Click on the image to play the video clip in a browser}.}
    \label{fig:ag2vidgtl}
\end{figure}

It turns out that humans do not perform perfectly in the above task. We quantified this effect in the following experiments on the SmthV2 dataset. We used the above annotation process for ground-truth videos (see ``Real'' row in Table~\ref{tab:semantic_table}). Interestingly, it can be seen from the reported accuracy in Table~\ref{tab:semantic_table} that our generated action videos of ``Move Down'' and ``Take'' are more easily recognizable by humans than the ground truth videos, perhaps because the latter might be a bit more diverse in motion than the predicted ones, which are learned to be very explicit. We observe similar behavior when looking at AG2Vid vs. AG2Vid-GTL synthesized videos, where the motion generated by AG2Vid is sometimes more explicit. For such an example, see the right most example of~\figgref{fig:ag2vidgtl}.

To evaluate the extent to which the AG2Vid model can generalize at test time to unseen actions, we manually defined four compositions of learned actions. In Table~\ref{tab:composing}, we show the semantic accuracy of the human evaluation we did for the new unseen actions: ``Huddle", ``Swap'', ``Right Up'' and ``Down Left''. For CATER new actions, the annotators were asked to choose the correct action from the of all possible CATER actions. For SmthV2, the annotators were asked to choose the right action from ``Right Up'' and ``Down Left''.

\begin{table}[ht]
    \centering
    \begin{adjustbox}{width=\columnwidth}

    \begin{tabular}{c|cccccccc}
    \hline
    \multirow{2}{*}{{Video Source}} & \multicolumn{8}{c}{{Standard Actions}}\\
    & Right & Up & Down & Left & Put & Take & Uncover & Cover \\ 
    \hline \hline

    Generated & 100. & 50. & 100. & 75. & 95. & 80. & 25. & 55. \\
    Real & 100. & 100. & 90. & 100. & 100. & 65. & 100. & 85. \\
    \hline 

    \end{tabular}
    
    \end{adjustbox}
    \caption{The semantic quality evaluated by humans of the generated and real action videos. We asked raters to select the action described in the video for each synthesized video with a given action. The table reports the accuracy of the human annotators with respect to the true action underlying the video. Actions above correspond to: 'Pushing [something] from left to right', 'Moving [something] up', 'Moving [something] down', 'Pushing [something] from right to left', 'Putting [something] on a surface', 'Taking [one of many similar things on the table]', 'Uncovering [something]', 'Covering [something] with [something]' .
}
    \label{tab:semantic_table} 

\end{table}

\subsection{Comparing AG2Vid to Scene-Graph Based Generation}
\begin{figure}[t]
    \centering
    \href{https://github.com/roeiherz/AG2Video/blob/master/Videos/Supp_Figure4.gif}{\includegraphics[width=\linewidth]{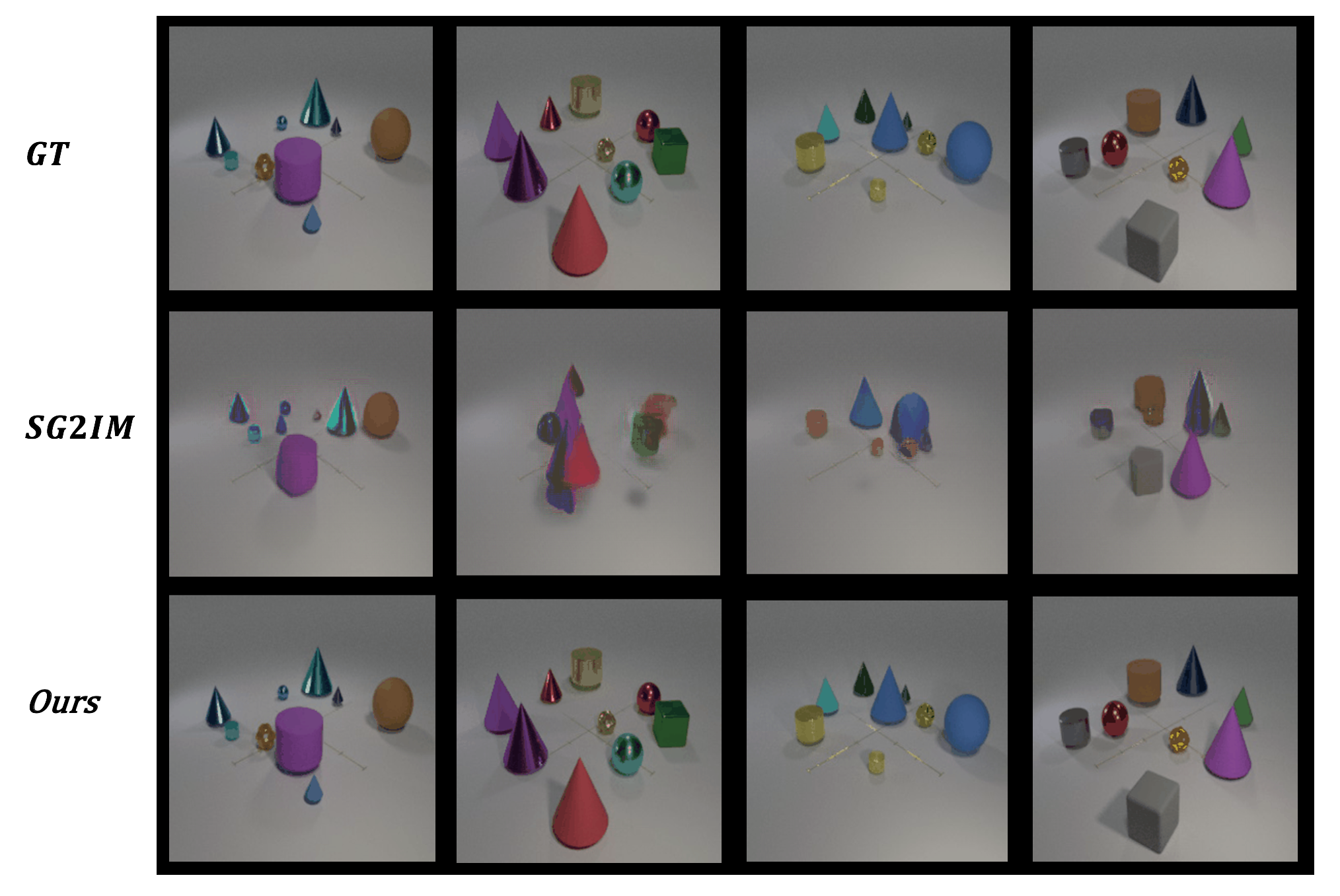}}
    \caption{Comparing Sg2Im and Ag2Vid results in CATER. Each column is a different sample. \textbf{Click on the image to play the video clip in a browser}.}
    \label{fig:cater_sg2img}
\end{figure}

Scene graphs are an expressive formalism for describing image content. Both datasets we use have frame-level scene graph annotation. Thus, for completeness we wanted to compare generation from these scene graphs with generation from action graphs. Towards this end, we used a scene-graph-to-image model~\citep{johnson2018image} trained to generate the images in the videos from their corresponding per frame scene graphs. This model does not condition the generation on actions, layout or the initial frame, and it only serves only for comparison in terms of realistic generation. It can be seen in Figure~\ref{fig:cater_sg2img} that the temporal coherency of AG2Vid is more consistent and coherent than the sequence of scene graphs, mainly because the Sg2im model is static and does not model the motion of objects.

\begin{figure}[t]
    \centering
    \href{https://github.com/roeiherz/AG2Video/blob/master/Videos/Supp_Figure5.gif}{\includegraphics[width=\linewidth]{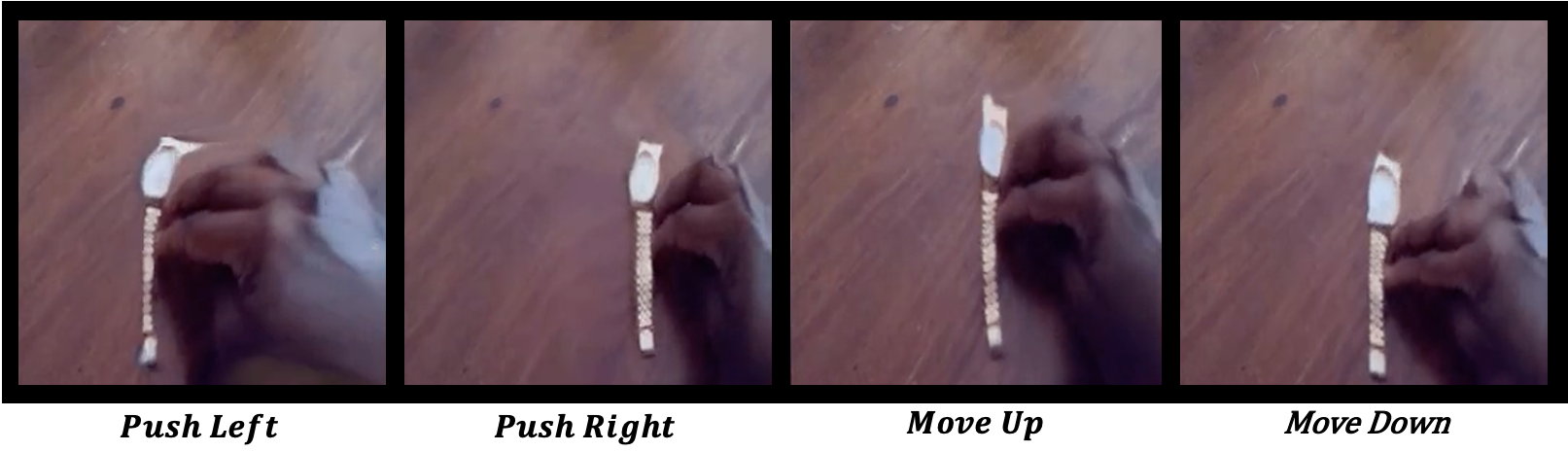}}
    \caption{Changing actions while fixing the first frame and layout over Something-Something V2 dataset. \textbf{Click on the image to play the video clip in a browser}.}
    \label{fig:using_ags}
\end{figure}

\subsection{Does the AG2Vid model take the Action Graph into account?}
To further demonstrate that this is not the case, and that the AG2Vid model indeed uses the AG, we provide more qualitative evidence in~\figgref{fig:using_ags}. To test this, we make changes to the AG while fixing the first frame and layout.

\end{document}